\documentclass[sigconf,authorversion]{aamas} 

\usepackage{algorithmic}
\usepackage[ruled,vlined,linesnumbered]{algorithm2e}
\usepackage[labelsep=colon]{caption}
\usepackage{subcaption}
\usepackage{csquotes}
\usepackage{footnote}
\usepackage{balance} 
\usepackage[mathscr]{euscript}

\DeclareSymbolFont{rsfs}{U}{rsfs}{m}{n}
\DeclareSymbolFontAlphabet{\mathscrsfs}{rsfs}

\setcopyright{ifaamas}
\acmConference[AAMAS '21]{Proc.\@ of the 20th International Conference on Autonomous Agents and Multiagent Systems (AAMAS 2021)}{May 3--7, 2021}{Online}{U.~Endriss, A.~Now\'{e}, F.~Dignum, A.~Lomuscio (eds.)}
\copyrightyear{2021}
\acmYear{2021}
\acmDOI{}
\acmPrice{}
\acmISBN{}

\title[AAMAS-2021 Formatting Instructions]{Self-Attention Meta-Learner for Continual Learning}

\author{Ghada Sokar}
\affiliation{
  \institution{Eindhoven University of Technology}
  \city{Eindhoven, The Netherlands}}
\email{g.a.z.n.sokar@tue.nl}

\author{Decebal Constantin Mocanu}
\affiliation{
  \institution{University of Twente \\ Eindhoven University of Technology}
  \city{The Netherlands}
  }
\email{d.c.mocanu@utwente.nl}

\author{Mykola Pechenizkiy}
\affiliation{
  \institution{Eindhoven University of Technology}
  \city{Eindhoven, The Netherlands}}
\email{m.pechenizkiy@tue.nl}

\begin{abstract}
Continual learning aims to provide intelligent agents capable of learning multiple tasks sequentially with neural networks. One of its main challenging, catastrophic forgetting, is caused by the neural networks non-optimal ability to learn in non-stationary distributions. In most settings of the current approaches, the agent starts from randomly initialized parameters and is optimized to master the current task regardless of the usefulness of the learned representation for future tasks. Moreover, each of the future tasks uses all the previously learned knowledge although parts of this knowledge might not be helpful for its learning. These cause interference among tasks, especially when the data of previous tasks is not accessible. In this paper, we propose a new method, named Self-Attention Meta-Learner (SAM)\footnote{Our code is available at https://github.com/GhadaSokar/Self-Attention-Meta-Learner-for-Continual-Learning}, which learns a prior knowledge for continual learning that permits learning a sequence of tasks, while avoiding catastrophic forgetting. SAM incorporates an attention mechanism that learns to select the particular relevant representation for each future task. Each task builds a specific representation branch on top of the selected knowledge, avoiding the interference between tasks. We evaluate the proposed method on the Split CIFAR-10/100 and Split MNIST benchmarks in the task agnostic inference. We empirically show that we can achieve a better performance than several state-of-the-art methods for continual learning by building on the top of selected representation learned by SAM. We also show the role of the meta-attention mechanism in boosting informative features corresponding to the input data and identifying the correct target in the task agnostic inference. Finally, we demonstrate that popular existing continual learning methods gain a performance boost when they adopt SAM as a starting point.
\end{abstract}


\newcommand{\BibTeX}{\rm B\kern-.05em{\sc i\kern-.025em b}\kern-.08em\TeX}

\begin{document}


\pagestyle{fancy}
\fancyhead{}


\maketitle 


\section{Introduction}
Lifelong learning aims to build machines that mimic human learning. The main characteristics of human learning are (1) humans never learn in isolation, (2) they build on the top of the learned knowledge in the past instead of learning from scratch, (3) and acquiring new knowledge does not lead to forgetting the past knowledge. These capabilities are crucial for autonomous agents interacting in the real world \cite{parisi2019continual,lesort2020continual}. For instance, systems like chatbots, recommendation systems, and autonomous driving interact with a dynamic and open environment and operate on non-stationary data. These systems are required to quickly adapt to new situations with the help of previous knowledge, acquire new experiences, and retain previously learned experiences. Deep neural networks (DNNs) have achieved outstanding performance in different areas such as visual recognition, natural language processing, and speech recognition \cite{zoph2018learning,chen2017deeplab,kenton2019bert,lin2017feature,guo2016deep,liu2017survey}. However, DNNs are very effective in domain-specific tasks (closed environments). Meanwhile, the performance degrades when the model interacts with non-stationary data, a phenomenon known as catastrophic forgetting \cite{mccloskey1989catastrophic}. Continual learning (CL) is a research area that addresses this problem and aims to provide neural networks with lifelong learning capability. 
\begin{figure*}[ht]
  \centering
  \includegraphics[width=\textwidth]{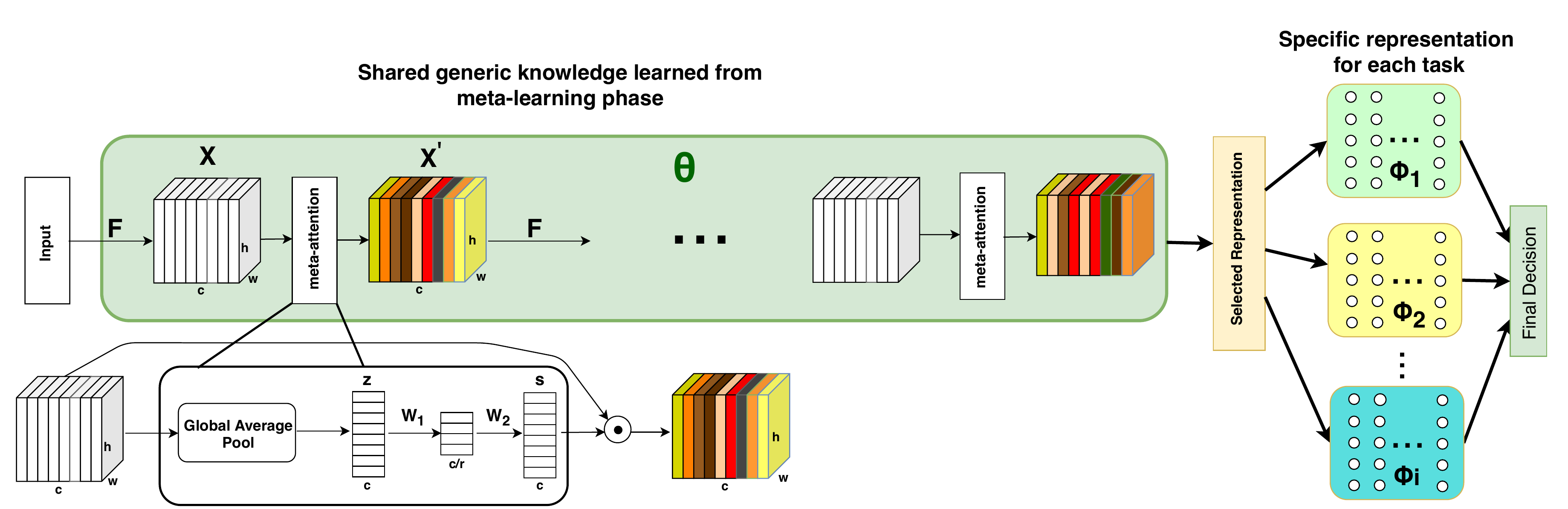}
  \caption{An overview of our proposed method, SAM. The network consists of two sub-networks. The first sub-network is trained using an optimization-based meta-learning algorithm to learn the prior generic knowledge $\theta$. This learned representation is shared between all tasks. A self-attention module is added after each layer to select the relevant representation for each task. The second sub-network contains a specific representation $\phi_{i}$ for each task $t_{i}$ that is learned on top of the selected representation whenever the model faces this task.}
  \label{method_overview}
\end{figure*}

Many works have been proposed to address the continual learning paradigm \cite{kirkpatrick2017overcoming,zenke2017continual,liu2018rotate,yoon2018lifelong,shin2017continual,sokar2021learning}. They can generally be categorized in regularization, architectural, and rehearsal approaches \cite{lange2019continual,parisi2019continual}. 
Another approach that recently showed its success in mitigating forgetting is the use of sparse connectivity \cite{mallya2018packnet,mallya2018piggyback} and/or sparse representations \cite{french1991using,sokar2020spacenet}. The continual learning paradigm targets many desiderata besides mitigating forgetting. The list of these desiderata includes but is not limited to: allowing forward and backward transfer, having a bounded system size, the inaccessibility of the previous tasks data, and the unavailability of the task identity during inference (see \cite{lesort2020continual,schwarz2018progress,diaz2018don} for the complete list). Since these desiderata are competing with each other, most of the previous methods target subsets of them. 

In this work, we shift the focus to some desiderata which are not widely addressed in the state-of-the-art to the best of our knowledge, and which are discussed in \cite{chen2018lifelong}. The first one is the necessity of having a good quantity of prior knowledge to help new tasks to learn in the continual learning paradigm. However, in most previous approaches, the model starts from randomly initialized parameters, then the parameters are optimized to achieve the highest performance on the first task. The knowledge gained from this task may contain only a bit or even no knowledge that is useful for future tasks \cite{chen2018lifelong}. Second, selecting the useful and relevant parts only from the previous knowledge to learn each of the future tasks instead of using the whole knowledge. We draw inspiration from \textit{human learning}. For instance, a computer science student should have a mathematical background to learn other advanced courses such as artificial intelligence, computer graphics, database management, simulation modeling, etc. This prerequisite knowledge facilitates learning each of these courses quickly. However, in each course, one picks only the relevant beneficial information from his/her mathematical background depending on the context of each course instead of using the whole knowledge. 
To address these desiderata, we propose to learn a prior representation for continual learning that permits and is more proficient at learning future tasks. We address the more realistic scenario where the autonomous agents might be deployed in an environment different from the ones they were pre-trained on. Therefore, we propose to learn this representation via meta-learning to permit generalization to out-of-distribution tasks. Moreover, instead of manually designing an algorithm for creating sparse representation, in this work, we take advantage of meta-learning to allow the network to learn to pick the \textit{relevant} sparse representation from the currently existing one, depending on the incoming data. To this purpose, we incorporate a self-attention mechanism with the meta-learner. At the continual learning time, we train each of the tasks in the sequence by building on the top of the selected sparse representation from the prior knowledge. These tasks are sampled from new distributions different from the one used in constructing the prior knowledge. Our empirical evaluation shows the importance of the proposed desiderata in the continual learning setting and their effectiveness in promoting the learning of each task and reducing the interference. We also demonstrate that building on the top of the relevant knowledge helps in identifying the correct target in the more challenging scenario where task identity is not available during inference (task agnostic scenario). 

Our contributions in this paper can be summarized as follows: First, we propose a Self-Attention Meta learner (SAM) that builds a prior knowledge that permits learning a continual sequence of tasks. In addition, SAM learns to pick the relevant representation for each of these tasks. Second, we address the more challenging and realistic scenario where the task identity is not available during inference (task agnostic). We also assume that the data of previous tasks is not accessible. Third, we achieved a better performance than the state-of-the-art methods by building on the top of the prior learned representation by SAM. Finally, we show that SAM significantly improves the performance of popular existing continual learning strategies.


\section{Continual Learning Formulation}

Before we describe the details of our proposed method, we introduce the problem formulation and our setting for continual learning. Continual learning problem consists of a sequence of $N$ tasks; $T_{CL}$= \{$t_{1}$, $t_{2}$,..., $t_{N}$\}. The tasks have non-stationary distributions. Each task $t_{i}$ consists of a set of samples, $D_{t_{i}}=\{(x^{1}_{i},y^{1}_{i}),(x^{2}_{i},y^{2}_{i}),...,(x^{n_{i}}_{i},y^{n_{i}}_{i})\}$,  where $n_{i}$ is the number of samples in task $t_{i}$. The data of each task are assumed to be sampled identically and independently (iid) from its corresponding distribution. All samples from the current task are observed before switching to the next task. Once the training of the current task ends, its data becomes not available. 

\textbf{Task agnostic inference} At any point in time, given the test input $x$, the learned model should predict the corresponding target from all classes learned so far regardless of the task identity. This is different from the common setting used in most of the previous methods for CL, task conditioned inference, where it is assumed that the test input contains a pair ($x$,$t_{i}$). However, the task identity $t_{i}$ is not always available during inference in real-world environments. In our default setting, we assume the unavailability of this information. 

\section{Self-Attention Meta-Learner (SAM)}
\label{method}
The ultimate goal of continual learning is to mimic human learning. The starting point for continual learning to address this goal is to collect and learn prior knowledge that can help in learning a sequence of tasks continuously. This prior knowledge should be characterized by the good generality that enables out-of-domain tasks to learn on top of it. Each task in the continual sequence should pick the relevant knowledge to it from the previously learned knowledge. In this section, we describe the details of our proposed method, SAM, that addresses these two goals. 

Figure \ref{method_overview} shows an overview of our proposed approach, SAM. The neural network consists of two parts. The first part represents the prior knowledge parameterized by the shared learned meta-parameters $\theta$. An attention module follows each layer in this shared sub-network which learns to pick the relevant features from that layer corresponding to the input. The second part learns specific representation to each task $t_{i}$ parameterized by $\phi_{i}$. Each task uses a few layers to capture the class specific discriminative features. The input to this part is the selected relevant knowledge from prior knowledge. At deployment time, the input $x$ is passed through the neural network f($x$;$\theta,\phi_{1},\phi_{2},..,\phi_{i}$,..) to predict the corresponding class from all learned classes so far.

We can divide our approach into two main phases: prior knowledge construction and the continuous learning of the tasks. The training procedure for these two phases is shown in Algorithm \ref{SAM_algo}. The details of each phase are discussed in the following paragraphs. 

\begin{algorithm}
\small
\caption{SAM}
\label{SAM_algo}
\begin{flushleft}
\textbf{Require}: $p$ ($\mathcal{T}_{meta}$): distribution over meta tasks

\textbf{Require}: $\alpha, \beta$: meta step size hyperparameters

\textbf{Require}: $\mathcal{N}_{meta}$: number of training steps for meta-learning

\textbf{Require}: $\eta$: step size hyperparameter for continual learning

\textbf{Require}: $\mathcal{N}_{CL}$: number of training steps for continual learning

\tcp{Learning prior knowledge with self-attention meta-learner}

Randomly initialize $\theta$ 

\For {n=1,...,$\mathcal{N}_{meta}$ steps}{
Sample batch of tasks $\mathcal{T}_{i}\sim p(\mathcal{T}_{meta})$

\For {all $\mathcal{T}_{i}$ }{
Sample minibatches $\mathscr{D}_{i}^{tr}$, $\mathscr{D}_{i}^{val}$ uniformly from $\mathscr{D}^{train}_{i}$

Forward pass the minibatch through the meta-learner including the attention modules using eq. \ref{squeeze}, \ref{excitation}, and \ref{recalibrate}

\tcp{One or few steps of gradient descent}
$\theta'_{i}=\theta - \alpha  \nabla_{\theta} \mathcal{L}(\mathscr{D}_{i}^{tr},\theta) $ 

}
$\theta=\theta - \beta  \nabla_{\theta} \sum_{\mathcal{T}_{i}} \mathcal{L}(\mathscr{D}_{i}^{val},\theta'_{i}) $ 

}
\tcp{Learning continuously a sequence of tasks ($t_{1},t_{2},t_{3},...)$}
Keep the meta-learned parameters $\theta$ fixed and shared for all tasks

\ForEach{$t_{i}$ in tasks ($t_{1}, t_{2},$ ...)}
{
Randomly initialize $\phi_{i}$ that represents the specific parameters to task $t_{i}$

\For {j=1,...,$\mathcal{N}_{CL}$ steps}{
Sample minibatch $D_{t_{i}}^{tr}$ from $D_{t_{i}}$

 $\phi_{i} =\phi_{i} - \eta \nabla_{\phi_{i}} \mathcal{L}(D_{t_{i}}^{tr},\phi_{i};\theta) $
}

}
\end{flushleft}
\end{algorithm}

\textbf{Prior knowledge construction.} As discussed earlier, the prior knowledge should generalize well to out-of-domain tasks. To satisfy this objective, we train the shared parameters $\theta$ using the optimization-based meta-learning algorithm MAML \cite{finn2017model} which proves its ability to generalize to out-of-distribution tasks \cite{finn2017meta}. MAML learns a parameter initialization $\theta$ that can quickly learn a new task using a small number of fine-tuning steps. In particular, the MAML algorithm consists of an \enquote{inner loop} (Algorithm \ref{SAM_algo}, Lines 4-7) and an \enquote{outer loop} (Algorithm \ref{SAM_algo}, Lines 2-8). In the inner loop, the parameters $\theta$ are adapted to multiple tasks using one or a few steps of gradient descent to obtain the parameters $\theta'_{i}$ which are specific for task instance ${\mathcal{T}_{i}}$. In the outer loop, the initialization $\theta$ is updated by differentiating through the inner loop to obtain a new initialization that improves inner-loop learning.  

We train our meta-learner using tasks ${\mathcal{T}_{i}} \sim p(\mathcal{T}_{meta})$ from a certain domain and optimizes the parameters such that when the model is faced with a new task, the model can adapt quickly. The objective of the MAML algorithm is:
\begin{equation}
\min_{\theta}  \sum_{{i}} \mathcal{L}(\theta-\alpha   \nabla_{\theta} \mathcal{L}(\theta,\mathscr{D}^{train}_{i}),\mathscr{D}^{test}_{i}), 
\end{equation}
where $\mathscr{D}^{train}_{i}$ corresponds to the training set for task $\mathcal{T}_{i}$ which is used in the inner optimization and $\mathscr{D}^{test}_{i}$ is the test data that is used for evaluating the outer loss $\mathcal{L}$. The inner optimization is performed via one or few steps of gradient descent with a step size $\alpha$. Further details for the meta-training procedure are included in Algorithm \ref{SAM_algo}. The $\mathcal{T}_{meta}$ tasks are used only for constructing the prior knowledge are different from the sequence of CL tasks that may come from another domain. 

\textbf{Selection of relevant knowledge}. The learned prior knowledge is shared between all tasks. When the model faces a new task, it builds on the top of this knowledge. Instead of using all the learned knowledge, the task picks the appropriate knowledge to use which helps in its learning. To address this point, we propose the self-attention meta-learner, SAM, by incorporating the self-attention mechanism proposed by \cite{hu2018squeeze} in our meta-learner. Rather than the standard training of the self-attention mechanism as in \cite{hu2018squeeze}, we make use of meta-learning to allow the network to learn to recalibrate the useful knowledge based on the input data. In particular, an attention block is added after each layer in the meta-learner (shared sub-network). The role of this block is to recalibrate the convolutional channels (or the hidden neurons) in each layer adaptively. It learns to boost the informative features corresponding to the input and suppress the less useful ones. The input of the attention block is the feature maps ${\textbf{X}} = \{X_{1}, X_{2},....,X_{c}\}$ resulted from applying the convolutional operator $F$ on the output of the previous layer, where $\textbf{X} \in \mathbb{R}^{h \times w \times c}$ and h, w, and c are the height, width, and depth of the feature maps. The output of each attention block is a vector {$ \textbf{s}$} of size $c$ that contains the rescaling value for each channel, where $c$ is the number of channels (depth). The recalibrated feature map $ {X}^{'}_{i} \in \mathbb{R}^{h \times w}$ is obtained as follows:
\begin{equation}
X^{'}_{i} = X_{i} \circ s_{i}, 
\label{recalibrate}
\end{equation}
where $s_{i}$ is the scalar value in the vector $\textbf{s}$ corresponding to the channel $i$ and the operation $\circ$ represents a channel-wise multiplication between the feature map $X_{i}$ and the scalar $s_{i}$.
The structure of the attention block is shown in Figure \ref{method_overview}. The attention and gating mechanism consists of two steps. The first step is to generate channel-wise statistics by compressing the global spatial information using global average pooling, resulting in a vector $\textbf{z}$ of $c$ channels. The $i$-th element of $\textbf{z}$ is calculated by: 
\begin{equation}
z_{i} = \frac{1}{h \times w} \sum_{k=1}^{h} \sum_{j=1}^{w}   X_{i}(k,j).
\label{squeeze}
\end{equation}
The second step is to model the interdependencies between channels. These dependencies are captured using two feedforward fully connected layers. The first layer is a bottleneck layer which reduces the dimension of the $c$ channels with a reduction ratio $r$ (which is a hyperparameter), followed by a second layer which increases the dimensionality back to $c$ channels. A sigmoid activation is applied to these $c$ channels as a simple gating mechanism, producing the output vector $\textbf{s}$ which can be calculated as follows: 

\begin{equation}
\textbf{s} = \sigma (\textbf{W}_{2} \delta(\textbf{W}_{1}\textbf{z})),
\label{excitation}
\end{equation}
where the operation $\delta$ is the ReLU function, $\sigma$ is the sigmoid function, and $\textbf{W}_{1}$ $ \in \mathbb{R}^{\frac{c}{r} \times c}$ and $\textbf{W}_{2}$ $ \in \mathbb{R}^{c \times \frac{c}{r}}$ are the parameters of the two feedforward fully connected layers respectively. 
The training of the attention mechanism is part of the meta-learning phase (e.g. $\textbf{W}_{1}$ and $\textbf{W}_{2}$ are part of the shared parameters $\theta$), therefore we called it \enquote{meta-attention}.

Previously, we illustrate the attention mechanism on convolutional neural networks. It is very easy to adapt it to multilayer perceptron networks by ignoring the global average pooling step. The recalibrated hidden neurons are generated by performing element-wise multiplication between the output vector of the attention block $\textbf{s}$ and the hidden neurons $\textbf{x}$.

\textbf{Learning a sequence of tasks.} When the model encounters a new task $t_{i}$, specific layers to this task parameterized by $\phi_{i}$ are added to the model. These layers can be any differential layers (e.g convolutional or feedforward fully connected layers). The input to these layers is the selected representation from the prior knowledge. The parameters $\phi_{i}$ are trained in an end-to-end manner with a specific output layer for this task with the following objective: 
\begin{equation}
\min_{\phi_{i}} \mathcal{L}_{i}(\phi_{i},D_{t_{i}};\theta),
\end{equation}
where $\mathcal{L}_{i}$ is the loss for task $t_{i}$ and  $D_{t_{i}}$ is the training set of task $t_{i}$. The shared parameters $\theta$ are kept fixed during learning each task.

\textbf{Final decision module}. This module predicts the final output $\hat{y}$ corresponding to the test input $x$ in the task agnostic scenario. First, we aggregate the output layers ($\hat{\textbf{y}}_{1}, \hat{\textbf{y}}_{2},...,\hat{\textbf{y}}_{N}$) of all tasks before applying the softmax, where $N$ is the number of encountered tasks so far. Then, the final output is predicted by getting the index of the highest value from the concatenated output vector. This index corresponds to the predicted class by the learned model and is calculated as follows:
\begin{equation}
 \arg \max_{idx} (\hat{\textbf{y}}_{1} \oplus \hat{\textbf{y}}_{2} \oplus... \oplus \hat{\textbf{y}}_{N}),
\label{final_decision}
\end{equation}
where $\hat{\textbf{y}}_{i}$ is the output layer of task $t_{i}$, $\oplus$ represents the concatenation operation over the output vectors, and $idx$ is the index of the element with the highest value.

\section{Experiments and Results}
\label{experiments}
We evaluate our method on the commonly used benchmarks for CL: Split CIFAR-10/100 and Split MNIST. We compare our method with state-of-the-art approaches in the regularization and architectural strategies. We also consider another baseline, \enquote{\textit{Scratch}}, where an independent network is trained for each task individually from scratch. Each independent network has the same architecture as the network trained across all CL tasks. To get the predicted class using this baseline in the task agnostic scenario, we are using the same final decision module as for SAM. We name this baseline in the Task Agnostic scenario as \enquote{\textit{Scratch(TA)}}. 
\subsection{Split CIFAR-10/100}
The Split CIFAR-10/100 benchmark \cite{zenke2017continual} consists of a combination of CIFAR-10 and CIFAR-100 datasets \cite{krizhevsky2009learning}. It contains 6 tasks. The first task contains the full dataset of CIFAR-10, while each subsequent task contains 10 consecutive classes from the CIFAR-100 dataset.

\subsubsection{Experimental Setup}
\label{cifar_architecture}
We follow the architecture used by Zenke et al \cite{zenke2017continual} and Maltoni et al. \cite{maltoni2019continuous} for a direct comparison with the baselines. The shared sub-network consists of 2 blocks. Each block contains a 3x3 convolutional layer followed by batch normalization \cite{ioffe2015batch}, a ReLU activation, and an attention module with a reduction ratio ($r$) equals to 8. The convolutional layer in each block has 32 feature maps. A max-pooling layer follows these two blocks. The shared sub-network is initialized by the self-attention meta-learner. The specific sub-network for each task consists of two 3x3 convolutional layers with 64 feature maps, followed by a max-pooling layer, one fully connected layer of a size 512, and an output layer. The specific layers for each task are randomly initialized. Each experiment is repeated 5 times with different random seeds. Additional training details are included in the appendix.

\textit{Self-attention meta-learner:} We trained a convolutional neural network on MiniImagenet \cite{ravi2016optimization}. The MiniImagenet dataset is a common benchmark used for few-shot learning. It contains 64 training classes, 12 validation classes, and 24 test classes. The model consists of 4 blocks with a max-pooling layer following each block. The number of feature maps in the convolutional layers in the first and second two blocks is 32 and 64 respectively. The learned weights of the first two blocks are copied to the shared sub-network as prior knowledge. 

Since there are differences between the structure of our architecture and the regular network used in the baselines, we analyze another split for the shared and specific sub-networks in our results and analysis, where the shared sub-network contains all layers except the output layer. In particular, the shared sub-network consists of all 4 blocks, initialized by the self-attention meta-learner, with a max-pooling layer follows every two blocks. The specific sub-network for each task contains only the output layer.

\subsubsection{Results}
\label{results_cifar}
\begin{figure}
  \centering
  \includegraphics[width=0.8\columnwidth]{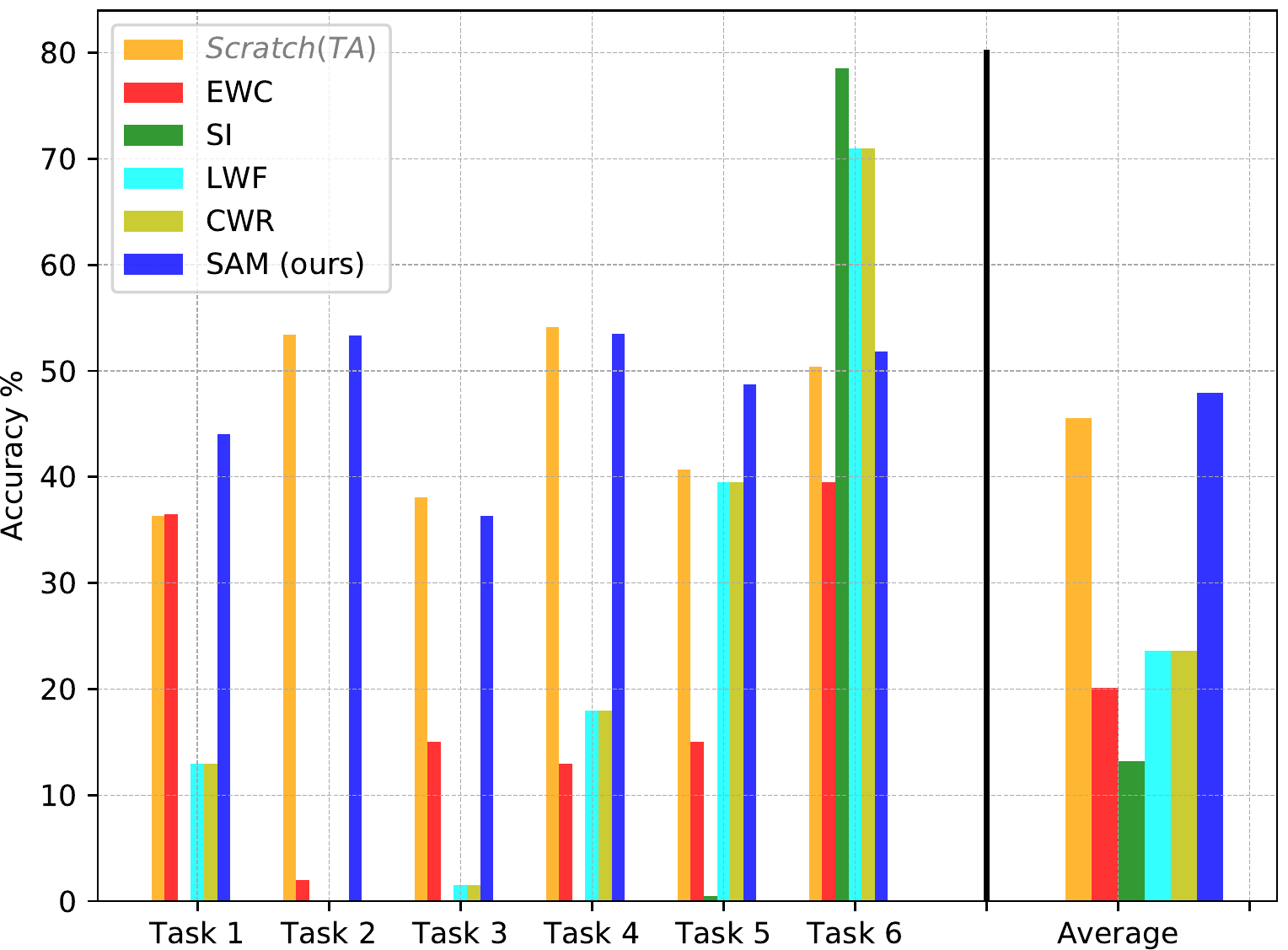}
  \caption{The accuracy of each task of the Split CIFAR-10/100 benchmark in the task agnostic inference after training all tasks. The \enquote{Average} x-axis label shows the average accuracy computed over all tasks for each method. Results for other methods except \enquote{Scratch(TA)} are reported from \cite{maltoni2019continuous}.}
  \label{cifar10_100_results}
\end{figure}
Figure \ref{cifar10_100_results} shows the accuracy of each task after training all tasks along with the average accuracy computed over all tasks. The regularization methods (EWC \cite{kirkpatrick2017overcoming}, LWF \cite{li2017learning}, and SI \cite{zenke2017continual}) suffer from forgetting the old tasks, while having a good performance on the last trained task. On the other hand, the performance achieved by SAM on each of the tasks is close to each other. SAM outperforms the regularization methods by a big margin. We also compare our method to the counterpart architectural baseline, CWR \cite{lomonaco2017core50}, which uses a fixed shared pre-trained knowledge for all tasks and trains an output layer for each task. As shown in the figure, the learned representation by SAM generalizes better than the CWR method. The average accuracy of our proposed method is higher than CWR by around 24.5\%. It worths to be highlighted that SAM achieves a performance that is better than optimizing a separate network for each task from scratch (\textit{Scratch(TA)}).

For a more fair comparison with the baselines, we perform another experiment where the shared sub-network consists of 4 blocks initialized by the self-attention meta-learner and a specific output layer for each task. Increasing the depth of the shared sub-network, while keeping it fixed, decreases the performance of SAM to 30.40\%. However, the prior knowledge learned by SAM is still more proficient, it performs better than the second-best performer method by around 7\%.

\subsection{Split MNIST}
Split MNIST \cite{zenke2017continual} is the most commonly used benchmark for CL. It consists of 5 tasks. Each task is to distinguish between two consecutive MNIST-digits. We study the performance of SAM and the baselines in the case of both, task agnostic and task conditioned inference.

\subsubsection{Experimental Setup}
\label{mnist_arch}
We use multilayer perceptron networks. Our model follows the same architecture used by Van et al. \cite{van2019three}. The shared sub-network consists of 2 blocks. Each block consists of one hidden layer with 400 neurons followed by a ReLU activation and an attention module with a reduction ratio ($r$) equals to 10. The shared sub-network is initialized by the self-attention meta-learner. A specific output layer for each task follows the shared sub-network. The weights of the output layer are randomly initialized. Since all the layers in the used architecture except the output are shared, we ensure a fair comparison with the baselines. Each experiment is repeated 5 times with different random seeds.

\textit{Self-attention meta-learner:} We trained a multilayer perceptron network on the Omniglot dataset \cite{lake2011one}. The Omniglot dataset consists of 1623 characters from 50 different alphabets. Each character has 20 instances. The images are resized to 28x28. The model has the same architecture of our shared sub-network, with a batch normalization that follows each hidden layer in each block. The learned weights of the model are used as prior knowledge. Additional training details are included in the appendix.

\subsubsection{Results} 
Table \ref{resultsSplitMNIST} shows the average accuracy over all tasks after training the CL sequence of tasks. We report the average accuracy in task agnostic as well as task conditioned inference.
As shown in the table, regularization methods achieve a very good performance in task conditioned inference, however, they suffer from huge accuracy drop in task agnostic scenario. SAM outperforms the regularization methods by more than 38\% in the latter case while achieving a comparable performance in the case of task conditioned inference. We also compare our approach to one of the well-known architectural approaches, DEN \cite{yoon2018lifelong}. DEN restores the drift in old tasks performance
using node duplication. The DEN method is originally evaluated on task conditioned scenario as the connections in this method are remarked with a timestamp (task identity) and in the inference, the task identity is required to test on the parameters that are trained up to this task identity only. We adapt the official code provided by the authors to evaluate the method in the task agnostic scenario and name it DEN(TA). After training all
tasks, we evaluate the test data on the model created each
timestamp. Then we used our proposed final decision module to get the final prediction. Similar to other baselines, SAM achieves a comparable performance to DEN in the task conditioned scenario. While the performance of SAM is better than DEN(TA) by 6\% in the task agnostic scenario. The DEN method expands each hidden layer by around 35 neurons, while SAM has a fixed number of parameters. 
Detailed behavior of SAM for CL in task agnostic inference is also included in the appendix.  

\begin{table}
  \caption{Average accuracy on the Split MNIST benchmark in case of task conditioned and task agnostic scenarios.}
  \label{resultsSplitMNIST}
  \centering
  \begin{tabular}{lcc}
    \toprule
    Method & Task conditioned & Task agnostic \\
    \midrule
    \textcolor{gray}{\textit{Scratch}} & \textcolor{gray}{\textit{99.68 $\pm$ \footnotesize{0.06}}} & - \\
    \textcolor{gray}{\textit{Scratch(TA)}} & - & \textcolor{gray}{\textit{67.8 $\pm$ \footnotesize{0.88}}}\\  
    EWC \cite{van2019three}& 98.64 $\pm$ \footnotesize{0.22} &  20.01$\pm$ \footnotesize{0.06} \\
    SI \cite{van2019three} & 99.09 $\pm$ \footnotesize{0.15} &19.99$\pm$ \footnotesize{0.06} \\
    LWF \cite{van2019three}&99.57 $\pm$ \footnotesize{0.02} & 23.85 $\pm$ \footnotesize{0.44}\\
    DEN & 99.26  $\pm$ \footnotesize{0.001} & -\\
    DEN(TA) & - & 56.95 $\pm$ \footnotesize{0.02}\\
    SAM (ours)& 97.95 $\pm$ \footnotesize{0.07} &  \textbf{62.63} $\pm$ \footnotesize{0.61}\\
    \bottomrule
  \end{tabular}
\end{table}
\section{Analysis}
\label{analysis}
In this section, we analyze: (1) the role of the meta-attention mechanism in the performance, (2) the selected representation by SAM for different classes, (3) the importance of having prior knowledge in the CL paradigm, and (4) the usefulness of the representation learned by SAM for learning tasks from a different domain.

\textbf{The role of the meta-attention mechanism in the performance}. We performed an ablation study to analyze the effect of the attention mechanism in the task agnostic scenario. We performed this experiment on the two studied benchmarks. For the Split CIFAR-10/100 benchmark, we analyzed the performance of the two studied splitting of the shared and specific sub-networks discussed in section \ref{cifar_architecture}. The results are summarized in Table \ref{AblationSplitMNIST}. Removing the meta-attention mechanism and using the whole knowledge in learning each task yield the lowest accuracy in all cases. Using an attention module at the last block only in the shared sub-network obtains a slightly better accuracy. The best performance is achieved using a meta-attention module in each block in the shared sub-network. We also observe that the importance of the attention mechanism increases when the shared sub-network gets deeper (Shared: 4 blocks) as the features become more discriminative at the higher layers. Feeding the relevant representation only from the prior knowledge to each specific classifier increases the confidence of the correct one, hence it helps in identifying the correct target.

\begin{figure}[ht]
 \centering
 \begin{subfigure}[b]{0.49\textwidth}
     \centering
     \includegraphics[width=0.79\textwidth]{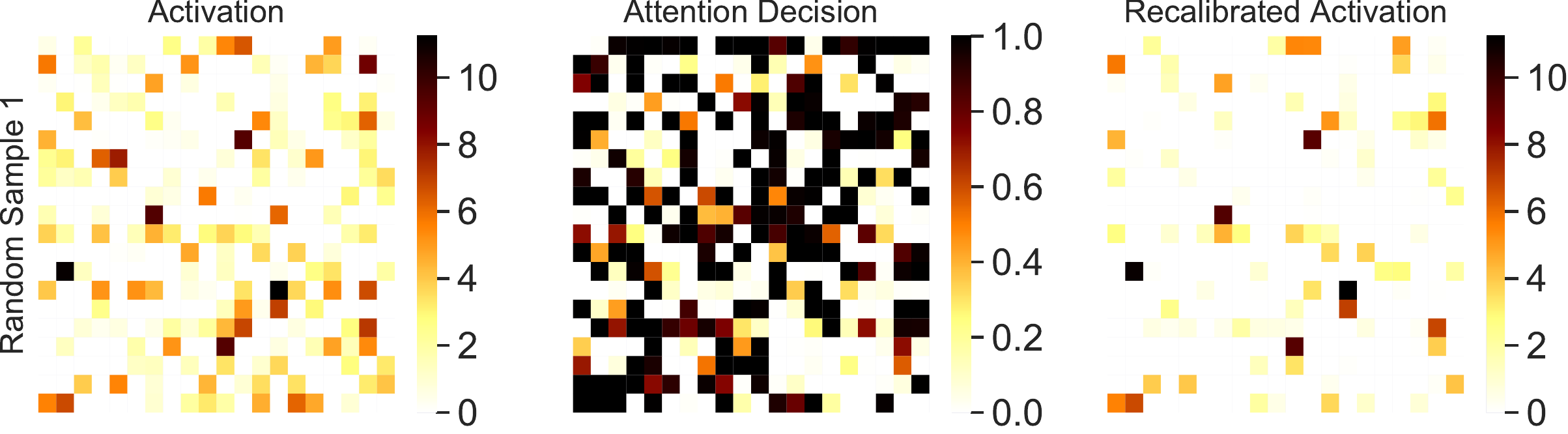}
     \includegraphics[width=0.79\textwidth]{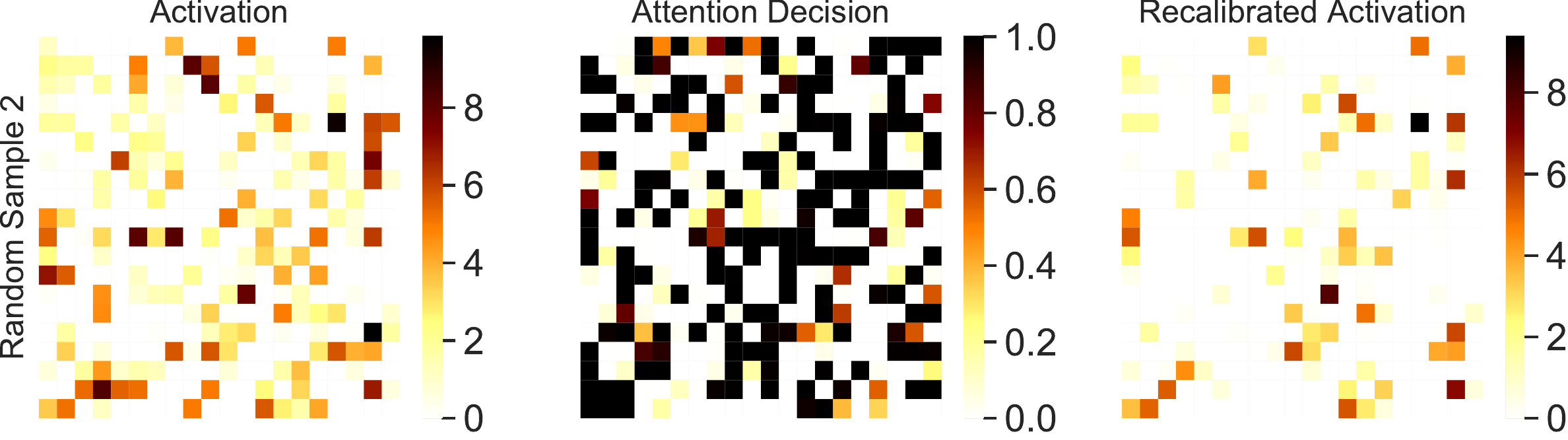}
     \caption{The first block}
     \label{Layer1}
\end{subfigure}
 \begin{subfigure}[b]{0.49\textwidth}
    \centering
    \includegraphics[width=0.79\textwidth]{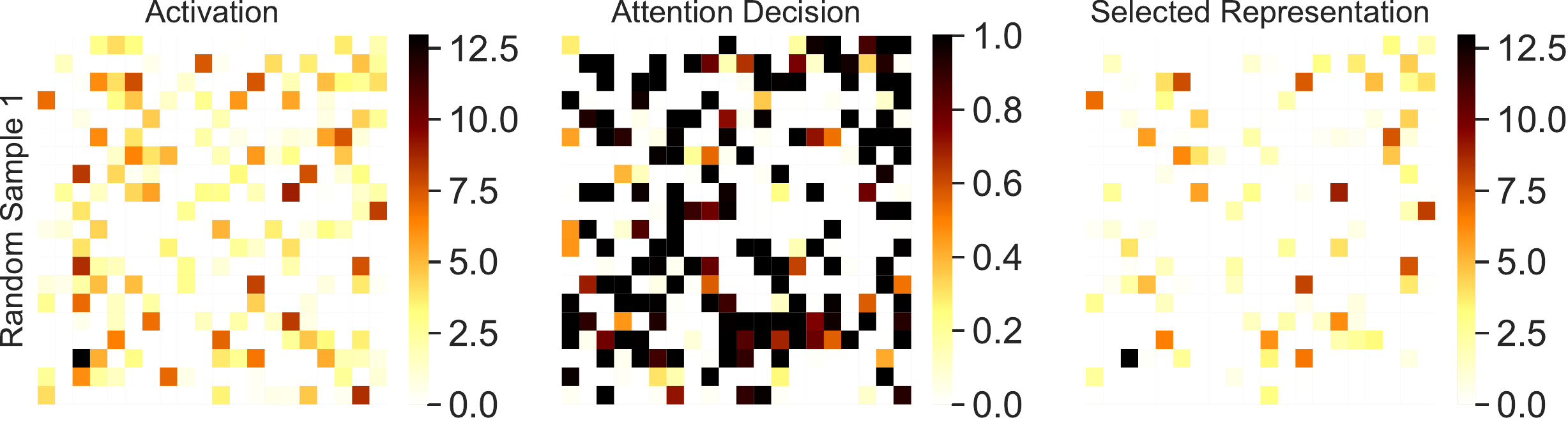}
    \includegraphics[width=0.79\textwidth]{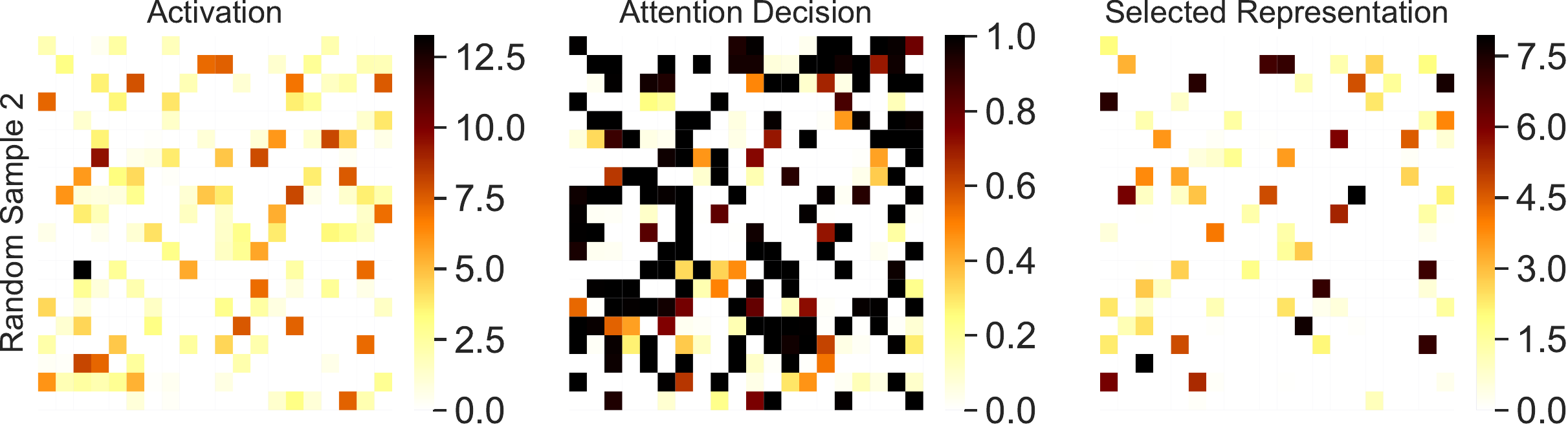}
      \caption{The second block}
     \label{Layer2}
 \end{subfigure}
 \caption{The visualization of the activation in shared sub-network for Split MNIST. The representation for each layer is reshaped to 20x20. Each row represents a random sample of a class in a certain task. Figures \ref{Layer1}  and \ref{Layer2} show the activation before being calibrated, the output of the attention module, and the recalibrated activation in the first and second blocks respectively. The last column represents the selected representation passed to the specific sub-networks.} 
 \label{Calibrated_Activation}
\end{figure}

\begin{table}
  \caption{Ablation study of SAM on the Split MNIST and Split CIFAR-10/100 benchmarks in the task agnostic scenario.}
  \label{AblationSplitMNIST}
  \centering
   \resizebox{\columnwidth}{!}{%
  \begin{tabular}{lccc}
  \toprule
         &Split MNIST& \multicolumn{2}{c}{Split CIFAR-10/100} \\
    \midrule
    Method & & Shared: 2 blocks &  Shared: 4 blocks\\
    \midrule
    SAM without attention & 51.33 $\pm$ \footnotesize{2.30}  & 46.52 $\pm$ \footnotesize{0.33} & 24.21 $\pm$ \footnotesize{0.59}\\
    SAM with attention at the last block only  & 54.67 $\pm$ \footnotesize{0.80} & {46.58 $\pm$ \footnotesize{0.29}} & 25.04 $\pm$ \footnotesize{0.14}\\
    SAM &  \textbf{62.63} $\pm$ \footnotesize{0.61} & \textbf{48.24} $\pm$ \footnotesize{0.30} & \textbf{30.40} $\pm$ \footnotesize{0.31}\\
    \bottomrule
  \end{tabular}
  }
\end{table}
\textbf{The selected representation}. In Figure \ref{Calibrated_Activation}, we draw the learned representation for each block in the shared sub-network before applying the attention module, the decision of the attention module, and the recalibrated activation. We draw these representations for random samples from different classes in different tasks from the Split MNIST benchmark. As shown in the figure, the relevant representation for each sample is sparse and it differs from one class to another. We also observe that more features are selected by the attention module in the first block where it is likely that different classes share many lower features with the prior knowledge. In the second block, the relevant features for the input are emphasized while many others are suppressed.

\begin{figure}
\begin{subfigure}[b]{\columnwidth}
  \centering
  \includegraphics[width=0.27\columnwidth]{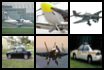}
   \caption{Task 1 (airplane, automobile)}
   \label{task_1}
  \end{subfigure}
  \begin{subfigure}[b]{0.49\columnwidth}
  \centering
  \includegraphics[width=0.50\columnwidth]{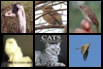}\\
    \includegraphics[width=0.90\columnwidth]{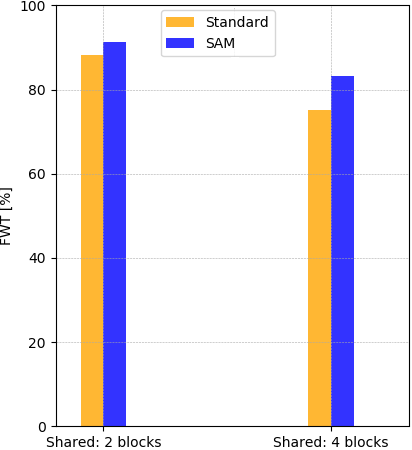}
   \caption{Task 2 (bird, cat)}
   \label{FWT_task_2}
  \end{subfigure}
    \begin{subfigure}[b]{0.49\columnwidth}
  \centering
    \includegraphics[width=0.50\columnwidth]{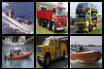}\\
    \includegraphics[width=0.87\columnwidth]{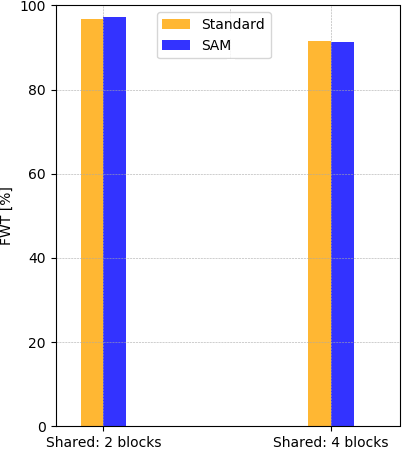}
   \caption{Task 3 (ship, truck)}
   \label{FWT_task_3}
  \end{subfigure}
  \caption{FWT comparison between SAM and the \enquote{Standard} setting used in most of the previous CL methods. In the Standard setting, the shared sub-network is initialized by the learned knowledge from Task 1 (airplane, automobile) from CIFAR10 benchmark (a). FWT is evaluated on a dissimilar task (bird, cat) (b) and another similar task (ship, truck) (c). SAM promotes the forward transfer in case of dissimilar tasks. While in the Standard setting, the knowledge learned by Task 1 contains less information useful for Task 2.}
  \label{FWT}
\end{figure}

\textbf{The prior knowledge}. We compare our method with the setting used in most of the previous continual learning methods, where the agent starts learning from randomly initialized parameters, maximizes the performance of the first task, and then faces the other tasks one by one. We call this setting \enquote{Standard}. We analyzed the forward transfer (FWT) in this setting as well as in SAM. FWT is a metric proposed by \cite{lopez2017gradient,diaz2018don} to assess the ability of the CL model to transfer knowledge for future tasks. We performed this analysis on the CIFAR10 dataset. We constructed three tasks from this dataset: two similar and one dissimilar. Each task contains two classes of CIFAR10. Task 1 contains the two classes of airplanes and automobiles. Task 2 contains bird and cat classes. While Task 3 contains ship and truck classes. We performed this analysis on the same architecture discussed in Section \ref{cifar_architecture} with the two different splittings of the shared and specific sub-networks. In the Standard setting, we use the same structure of the SAM network. The shared sub-network in this setting is initialized with the weights learned from training the model on Task 1 (airplane, automobile). While in SAM, the shared sub-network is initialized by the self-attention meta-learner reported in Section \ref{cifar_architecture}. The shared sub-network is then frozen in both cases. We then evaluate the FWT on Task 2 and Task 3. We adapted the FWT metric to compare the two settings. The FWT is estimated by the accuracy obtained on each task using the fixed shared sub-network, while allowing the training of the specific sub-network. As shown in Figure \ref{FWT_task_3}, the forward transfer in the case of the Standard setting is close to SAM for Task 3 (ship, truck), where Task 1 contains useful knowledge for Task 3 as they have some similarities. While for Task 2 (bird, cat), the FWT of SAM is higher than the Standard setting as shown in Figure \ref{FWT_task_2}. Moreover, the figure shows that this gap in performance increases when the shared sub-network gets deeper. The FWT of SAM is higher than the Standard setting by around 8\%. The experiment reveals the importance of having a good quantity of prior knowledge for the CL paradigm to promote learning future tasks.

\begin{figure}[ht]
 \centering
 \begin{subfigure}[b]{0.49\textwidth}
     \centering
     \includegraphics[width=0.8\textwidth]{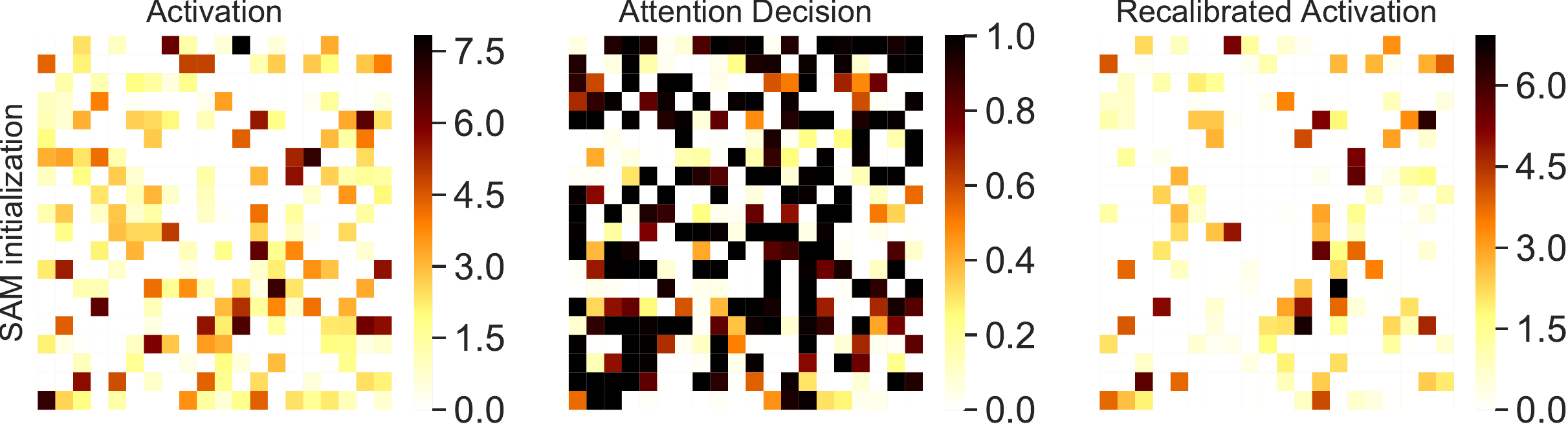}
     \includegraphics[width=0.8\textwidth]{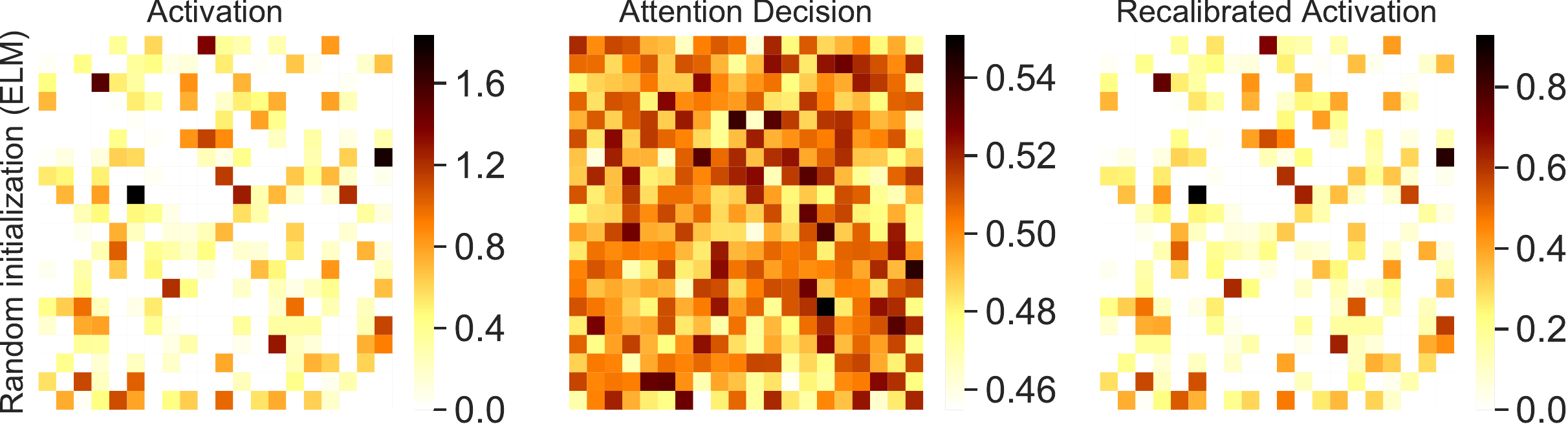}
     \caption{The first block}
     \label{R_Layer1}
\end{subfigure}
 \begin{subfigure}[b]{0.49\textwidth}
    \centering
    \includegraphics[width=0.8\textwidth]{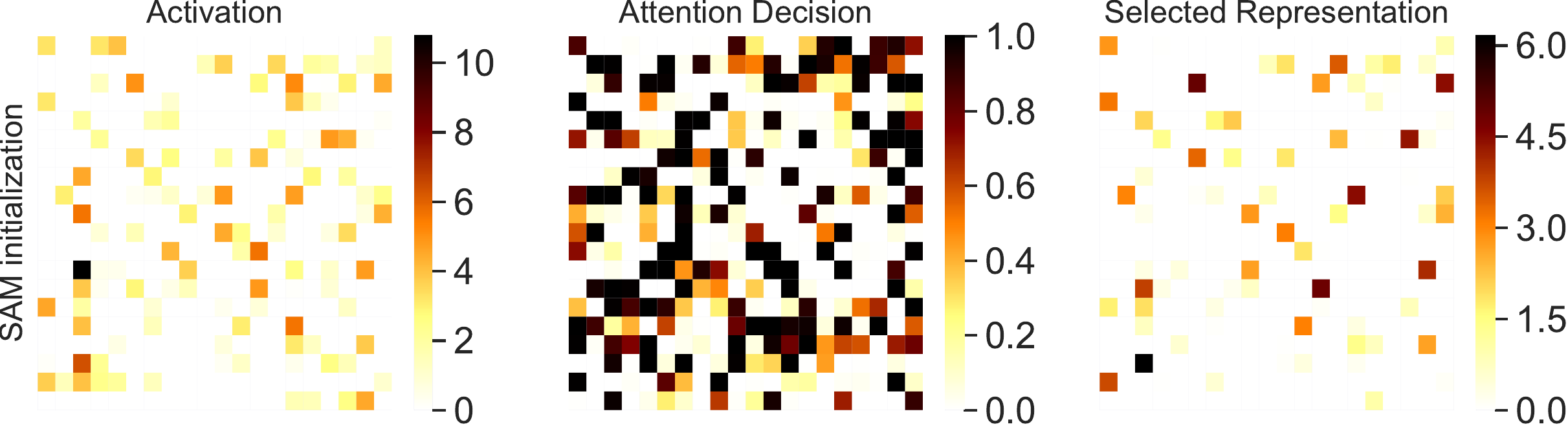}
    \includegraphics[width=0.8\textwidth]{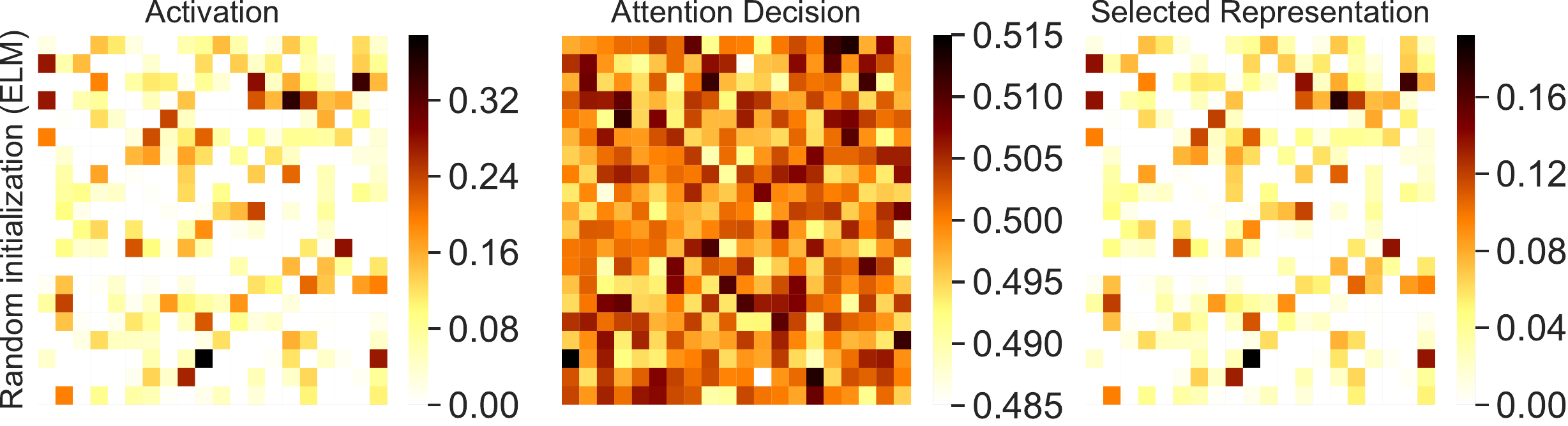}
      \caption{The second block}
     \label{R_Layer2}
 \end{subfigure}
 \caption{Activations in the shared sub-network for the Split MNIST benchmark. The representation is reshaped to 20x20. The first and second rows in show the representations when the shared sub-network is initialized by SAM and a random initialization (ELM) respectively.}
 \label{Compare_Random_Calibrated_Activation}
\end{figure}

\textbf{Is the representation learned by the self-attention meta-learner on a certain domain useful for learning a sequence of tasks from another domain?} To answer this question, we compare SAM to the extreme learning machine (ELM) which was first proposed by Huang et al. \cite{huang2004extreme} and which we adapted to fit our CL settings. They proposed the \enquote{generalized} networks that provide the best generalization performance at fast learning speed. The network parameters are randomly initialized and are not updated, while the parameters of the output layer are learned. This research is then extended by many works \cite{barak2013sparseness,rigotti2013importance,fusi2016neurons,bai2015generic,huang2015local}. We compare SAM with ELM by initializing the shared sub-network randomly and keeping its weights fixed. We visualize the representation of each hidden layer in the shared sub-network on the Split MNIST benchmark in case of initializing the weights by the self-attention meta-learner and in case of randomly initialized weights (ELM). As shown in Figure \ref{Compare_Random_Calibrated_Activation}, the self-attention meta-learner boosts and selects some features, while the random initialization gives equal importance for each feature. Table \ref{randomvsmeta} shows a comparison between the two methods in terms of performance on the two studied benchmarks. As shown in the table the prior knowledge learned by SAM generalizes better for the CL tasks despite that the domain is different. 

\begin{table}
  \caption{A comparison between the random initialization (ELM) and the initialization by the generic prior knowledge (SAM) for the shared sub-network.}
  \label{randomvsmeta}
  \centering
  \begin{tabular}{lcc}
    \toprule
    Method  & Split CIFAR-10/100 & Split MNIST\\
    \midrule
    ELM &  37.95 $\pm$ \footnotesize{0.64} & 58.42 $\pm$ \footnotesize{0.91} \\  
    SAM (ours) & \textbf{48.24} $\pm$ \footnotesize{0.30} & \textbf{62.63} $\pm$ \footnotesize{0.61} \\
    \bottomrule
  \end{tabular}
\end{table}

\section{Improvements of state-of-the-art CL approaches with SAM}
We have shown the effectiveness of the prior knowledge learned by SAM for continual learning. The experimental evaluation demonstrates that SAM achieves better performance than the state-of-the-art methods by quickly adapting a few layers on the top of the learned representations. Yet, accumulating the knowledge from each task is one of the desiderata for the CL paradigm. In this section, we analyze the performance of popular CL methods when they are enhanced with SAM. We evaluate the performance of the regularization method SI \cite{zenke2017continual} as well as the optimization-based meta-learning method MER \cite{riemer2018learning} when they are combined with SAM as well as their original form. In particular, instead of freezing the shared sub-network after learning the prior knowledge, we allow for accumulating the knowledge from each CL task. The shared sub-network is updated using a CL baseline (SI or MER). Accumulating the knowledge from each task in the shared sub-network causes catastrophic forgetting to the previously learned ones. The SI method addresses this problem by adding a regularization term in the loss function to constrain the change in the important weights of previous tasks. In the MER method, a small memory for experience replay is used and the parameters are trained using the Reptile meta-learning algorithm \cite{nichol2018reptile}. We also add the simple fine-tuning method as another baseline, where new tasks are trained continuously without any mechanism to avoid forgetting in the shared sub-network. We perform this experiment on the Split MNIST and Split CIFAR-10/100 benchmarks. We use the same architectures and training details used for evaluating our method (Section \ref{experiments}). To ensure a fair comparison with the original form of the methods, for the Split CIFAR-10/100 benchmark, we use the split of the network where the shared sub-network consists of 4 blocks, while each specific sub-network contains only an output layer. For the MER algorithm, we adapt the official code provided by the authors to evaluate it on the Split MNIST benchmark. Following the notation of their paper, we use a batch size (k-1) of 10, the number of batches per example of 5, $\gamma$ of 1.0, and $\beta$ of 0.01. We use a memory buffer of size 200 to learn 1000 sampled examples across each task. The results are shown in Table \ref{combineSAMwithothers}. 
\begin{savenotes}
\begin{table}
  \caption{Enhancing existing continual learning strategies by SAM. \enquote{Standard} represents the original form of the methods. The accuracy is reported on the Split MNIST and Split CIFAR-10/100 benchmarks in the task agnostic scenario.}
  \label{combineSAMwithothers}
  \resizebox{\columnwidth}{!}{%
  \centering
  \begin{tabular}{lcccc}
    \toprule
        &\multicolumn{2}{c}{Split MNIST}& \multicolumn{2}{c}{Split CIFAR-10/100}\\
    \midrule
    Method  & Standard & SAM & Standard & SAM \\
    \midrule
    Fine-tuning & 19.86 $\pm$ \footnotesize{0.04} & \textbf{53.87} $\pm$ \footnotesize{1.73} &12.24 $\pm$ \footnotesize{0.05}& \textbf{25.45} $\pm$ \footnotesize{1.76}\\  
    SI & 19.99 $\pm$ \footnotesize{0.06} & \textbf{67.32} $\pm$ \footnotesize{0.43} &13.39 $\pm$ \footnotesize{0.04} & \textbf{42.92} $\pm$ \footnotesize{1.01}\\ 
    MER & 32.66 $\pm$ \footnotesize{2.33}  & \textbf{50.04} $\pm$ \footnotesize{1.85} & -\footnote{As the original paper evaluated the method on multi-layer perception networks only, we restrict our evaluation here on the Split MNIST dataset.} & -\\
    \bottomrule 
  \end{tabular}
  }
\end{table}
\end{savenotes}

Interestingly, when SAM is combined with the other methods, it always improves their performance. Moreover, the combination of SAM with the fine-tuning baseline increases its performance despite that there is no forgetting avoidance strategy. MER achieves good results despite of using only 1000 samples from each task. Although the regularization methods suffer from a huge performance drop when applied in the task agnostic scenario as shown before, combining SAM with the regularization method (SI) leads to a significant improvement: around 47\% and 29\% on the Split MNIST and Split CIFAR-10/100 benchmarks respectively. SAM reduces the forgetting by allowing an adaptive update for the weights. The update of the weights becomes a function of the recalibrated activations by SAM. Therefore, the knowledge accumulated by the new tasks affects a subset of the previously learned representation. Accumulating the knowledge in SAM while using the SI method to constrain the change in the important weights of old tasks outperforms SAM alone by 5\% and 13\% on the Split MNIST and Split CIFAR-10/100 benchmarks respectively. This analysis assures the importance of our proposed desiderata and method for continual learning. We believe that it would open up many directions for the task agnostic scenario by adopting SAM as the starting point.  

\section{Related Work}
The idea of lifelong learning dates back to the 1990s \cite{thrun1998lifelong,thrun1995lifelong}. Recently this learning paradigm received a lot of attention. Many works have been proposed to address the catastrophic forgetting issue \cite{mcclelland1995there,mccloskey1989catastrophic} in neural networks. Regularization approaches add a regularization term to the learning objective to constrain the changes in important weights for the past learned tasks \cite{aljundi2018memory,kirkpatrick2017overcoming,zenke2017continual}. The way of estimating the parameter importance differs between these approaches. In Elastic Weight Consolidation (EWC) \cite{kirkpatrick2017overcoming}, weight importance is calculated using an approximation of the diagonal of Fisher information matrix. In Synaptic Intelligence (SI) \cite{zenke2017continual}, the importance of the weights is computed in an online manner during training. The importance is estimated by the amount of change in the loss by a weight summed over its trajectory. On the other hand, in Memory Aware Synapses (MAS) \cite{aljundi2018memory} the weights are estimated using the sensitivity of the learned function rather than the loss. Learning Without Forgetting (LWF) \cite{li2017learning} is another regularization method that constrains the change of model predictions on the old tasks, rather than the weights, by using a distillation loss \cite{hinton2015distilling}. Another approach is to replay the data of the old tasks along with each task learning. Gradient Episodic Memory (GEM) and iCaRL \cite{lopez2017gradient,rebuffi2017icarl} methods keep a subset of the old tasks data and combine the replay with regularization term in the objective. The Generative Replay method proposed in 2016 in \cite{mocanu2016online} and revised in 2017 in \cite{shin2017continual} keeps a generative model to generate data for the previous tasks instead of storing the original one. Many other works have been proposed afterward based on this idea \cite{van2020brain,sokar2021learning}. Other existing techniques modify the model architecture to adapt to a sequence of tasks. Progressive Neural Network (PNN) \cite{rusu2016progressive} instantiates a new neural network for each task and keeps previously learned networks frozen. CopyWeights with Reinit (CWR) \cite{lomonaco2017core50}, is a counterpart to PNN, where a fixed number of shared parameters is used for all tasks. The shared knowledge comes from freezing the shared weights after training the first batch. They initialize the shared weights using random weights or from a pre-trained model (ImageNet). Dynamic expandable network (DEN) \cite{yoon2018lifelong} expands the model when the performance of old tasks decreased. Other methods use sparse connections and keep a mask for each task to leave space for other tasks \cite{mallya2018piggyback,mallya2018packnet}. While they require the availability of the task identity during inference to select the corresponding mask. Recent work by Sokar et al. \cite{sokar2020spacenet} proposed a sparse learning algorithm, SpaceNet, to produce sparse representation to mitigate forgetting. In most of the previous methods, the agent starts learning from randomly initialized parameters, facing the sequential tasks one by one. Each task builds on all the previously learned knowledge.

Attention has emerged as an improvement in machine translation systems in natural language processing by \cite{bahdanau2014neural}. Recently, attention mechanisms have been addressed in many computer vision tasks \cite{chen2017sca,huang2019attention,wang2017residual,fu2019dual,zhang2018context,hu2018squeeze,parmar2019stand}. Few works have used the attention mechanisms in the CL paradigm. Serra et al. \cite{serra2018overcoming} proposed attention distillation loss combined with the distillation loss from \cite{li2017learning} to constrain the changes in the old tasks. Dhar et al. \cite{dhar2019learning} proposed a hard attention mechanism that determines the important neurons for each task. These units are masked during learning future tasks.

Very recently in 2019, a new direction arises by combining meta-learning methods with the CL paradigm. Finn et al. \cite{finn2019online} proposed modification for the MAML algorithm to work for an online setting. They focus on maximizing the forward transfer and sidestep the problem of catastrophic forgetting by maintaining a buffer of all the observed data. MER \cite{riemer2018learning} combines experience replay with the Reptile \cite{nichol2018reptile} meta-learning algorithm in an online setting. Instead of storing all the observed data, they keep a fixed size memory for all tasks and update the buffer with reservoir sampling. Another approach by \cite{javed2019meta} uses the meta-learning paradigm to learn to continually learn. They pre-train the network and continually learn tasks sampled from the same distribution. Caccia et al. \cite{caccia2020online} aim to target the more realistic scenario, where the continual tasks may come from new distribution not encountered during pre-training. However, they relax the assumptions by allowing for task revisiting and optimize for fast adapting. The setting of the CL problem differs in each of these methods. Yet, meta-learning seems a promising direction for solving the CL problem in all these different settings. 
\section{Conclusion}
In this paper, we address two desiderata of the continual learning paradigm that help deep neural networks in learning continuously a sequence of tasks and which are largely overlooked in state-of-the-art. First, the necessity of having a good quantity of prior knowledge to promote future learning. Second, selecting the relevant knowledge to the current task from the previous knowledge instead of using the whole knowledge. To this purpose, we propose SAM, a self-attention meta-learner for the continual learning paradigm. SAM learns a prior knowledge that can generalize to new distributions and learns to boost the features relevant to the input data. During the continual learning phase, we introduce out-of-domain tasks. Our empirical evaluation and analysis show the effective role of our proposed desiderata in improving the performance of the CL paradigm. The experimental results show that the proposed method outperforms the state-of-the-art methods from different continual learning strategies in the task agnostic inference by a large margin: at least 25\% and 5.6\% on the Split CIFAR-10/100 and Split MNIST benchmarks receptively. Remarkably, SAM achieved performance on par with the Scratch(TA) baseline despite that in this baseline an independent model is optimized for each task separately. Finally, we demonstrate that combining SAM with the existing continual learning methods boosts their performance. Our results suggest that continual learning has the potential of outperforming typical single task learners on classification accuracy in the task agnostic scenario. This may prove to be the turning point of a change in paradigm in the neural networks exploitation style, enabling novel benefits with neural networks, and opening the path for several new research directions.

\balance
\bibliographystyle{ACM-Reference-Format} 
\bibliography{sample}

\newpage
\clearpage
\appendix

\section{Additional Experiment Details}
\subsection{Continual Learning Training Details}
\label{training_details}
On the Split MNIST benchmark, each task is trained for 5 epochs. We use a batch size of 64. The model is trained using stochastic gradient descent with Nesterov momentum of 0.9 and a learning rate of 0.01. We search in the space \{2,4,8,10,20\} for the reduction ratio ($r$). 

The standard training/test-split for the MNIST dataset was used, resulting in 60,000 training images and 10,000 test images.

On the Split CIFAR-10/100 benchmark, each task is trained for 30 epochs. We use a batch size of 64. The model is trained using Adam optimizer ($\eta  = .001, \beta1 = 0.9, \beta2 = 0.999)$.  We search in the space \{2,4,8,16\} for the reduction ratio ($r$). The CIFAR-10 dataset consists of 10 classes and has 60000 samples (50000 training + 10000 test), with 6000 images per class. While the CIFAR-100 dataset contains 600 images per class (500 train + 100 test).

Each experiment is repeated 5 times with different random seeds.
\subsection{Meta-training Details}
\label{meta_training_details}
On the MiniImagenet dataset, the self-attention meta-learner was trained using 5-way classification, 5 shot training, a meta batch-size of 4 tasks. The model was trained using 5 gradient steps with step size $\alpha$ = 0.01. The meta step size for the outer loop $\beta$ = 0.001. The model was trained for 60000 iterations.

On the Omniglot dataset, the self-attention meta-learner models were trained using 5-way classification, 1 shot training, and a meta batch-size of 32 tasks. The models were trained using 5 inner gradient steps with step size $\alpha$ = 0.4. The meta step size for the outer loop $\beta$ = 0.001. The models were trained for 1000 iterations.

We adapt the Pytorch code for the MAML algorithm from \cite{MAML_Pytorch} to incorporate the attention module in the network architecture and to work for multilayer perceptron networks.

\section{Gradual Learning Behavior of SAM}
\label{detailed_behavior}
\begin{figure}[ht]
 \centering
 \includegraphics[width=0.48\columnwidth]{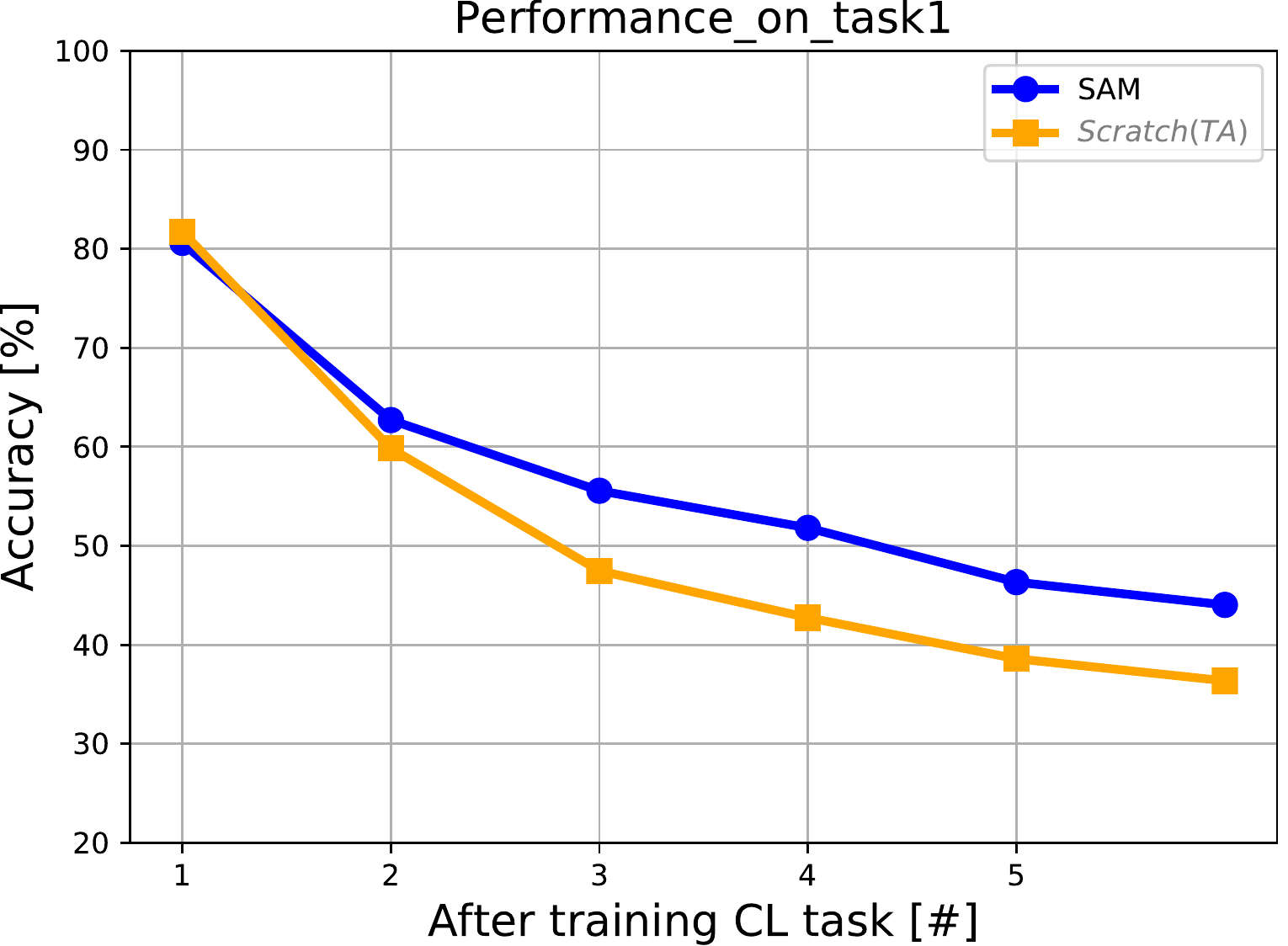}
 \includegraphics[width=0.48\columnwidth]{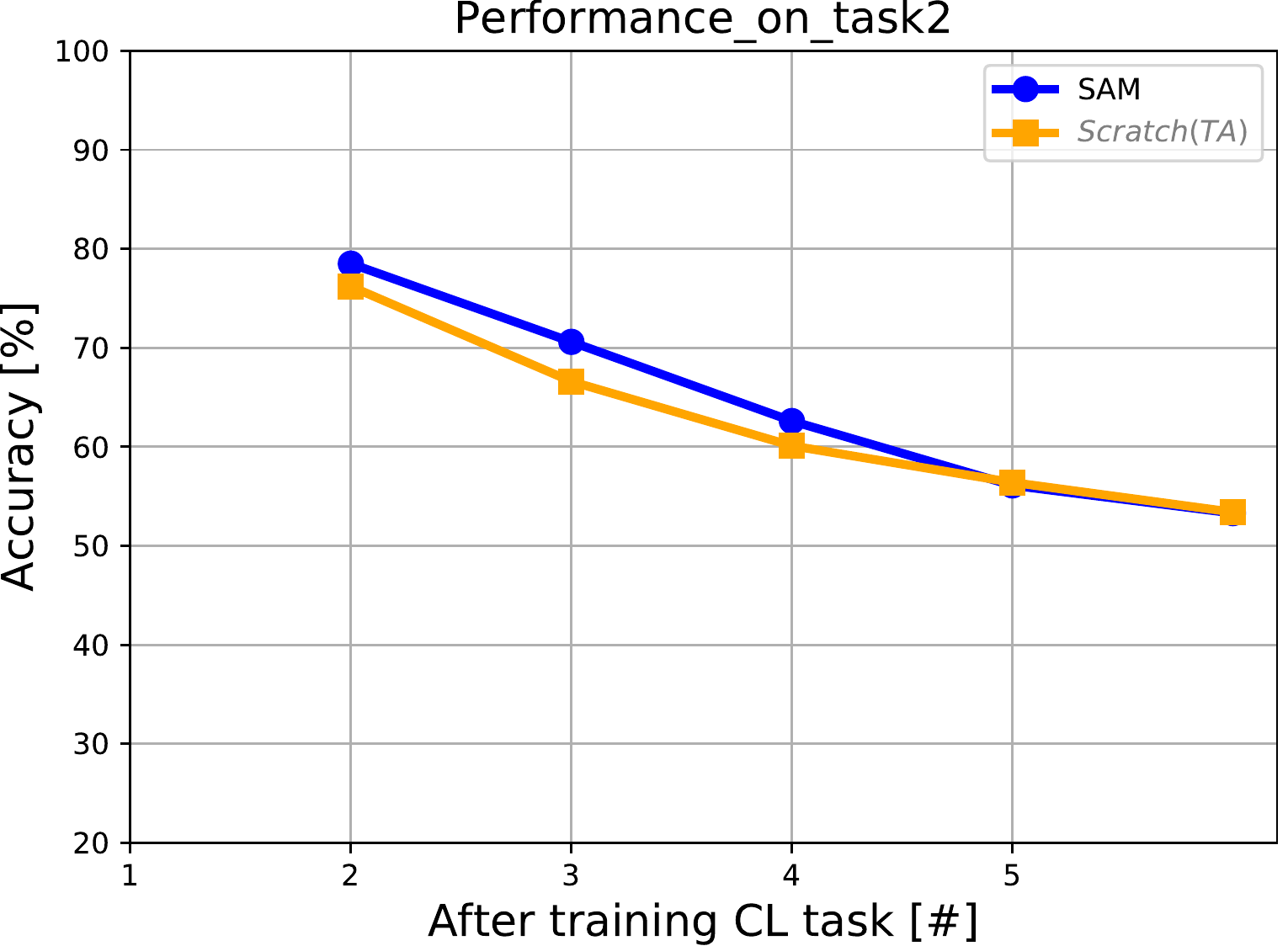}\\
  \vspace{0.4cm}
 \includegraphics[width=0.48\columnwidth]{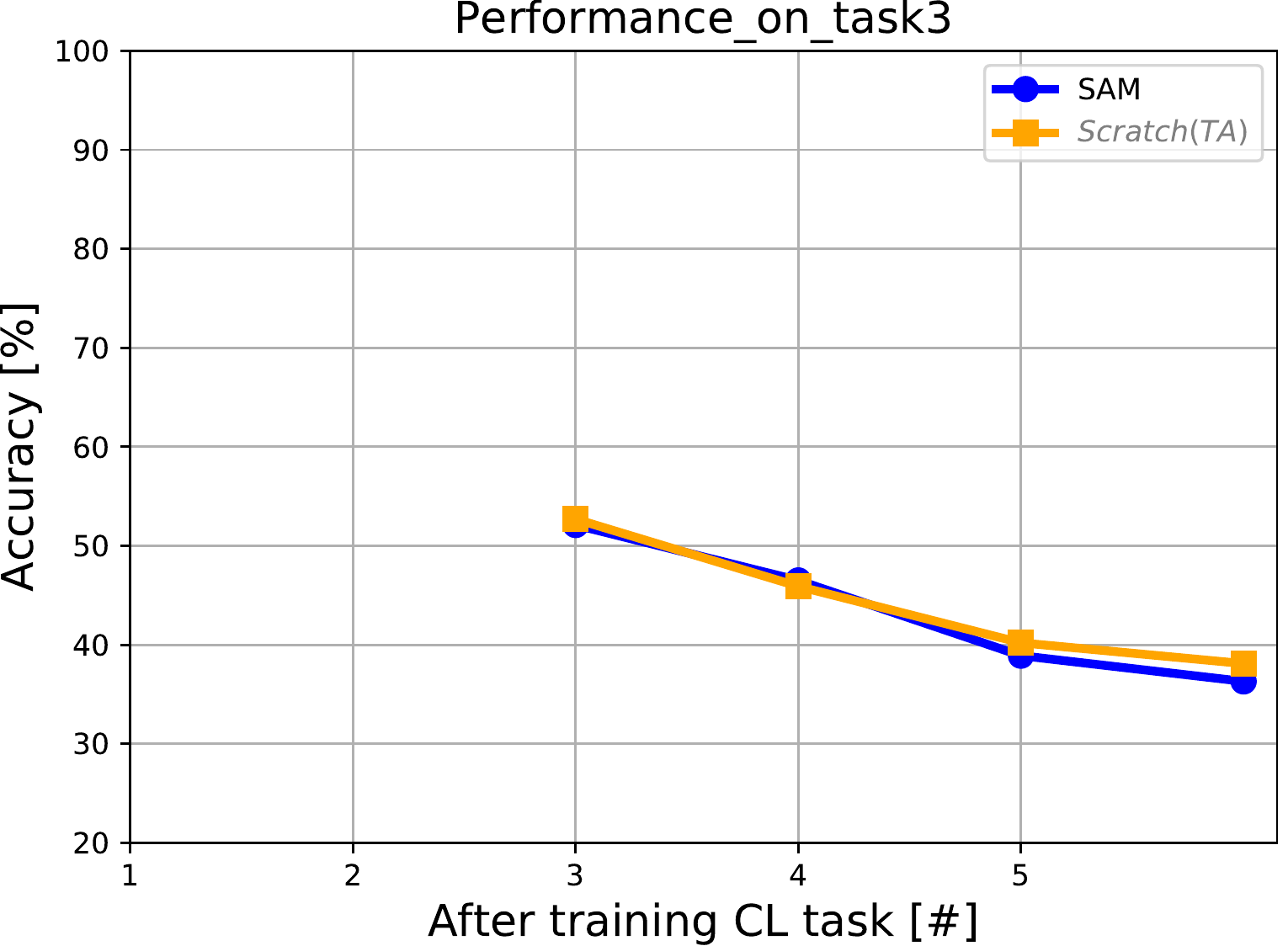}
 \includegraphics[width=0.48\columnwidth]{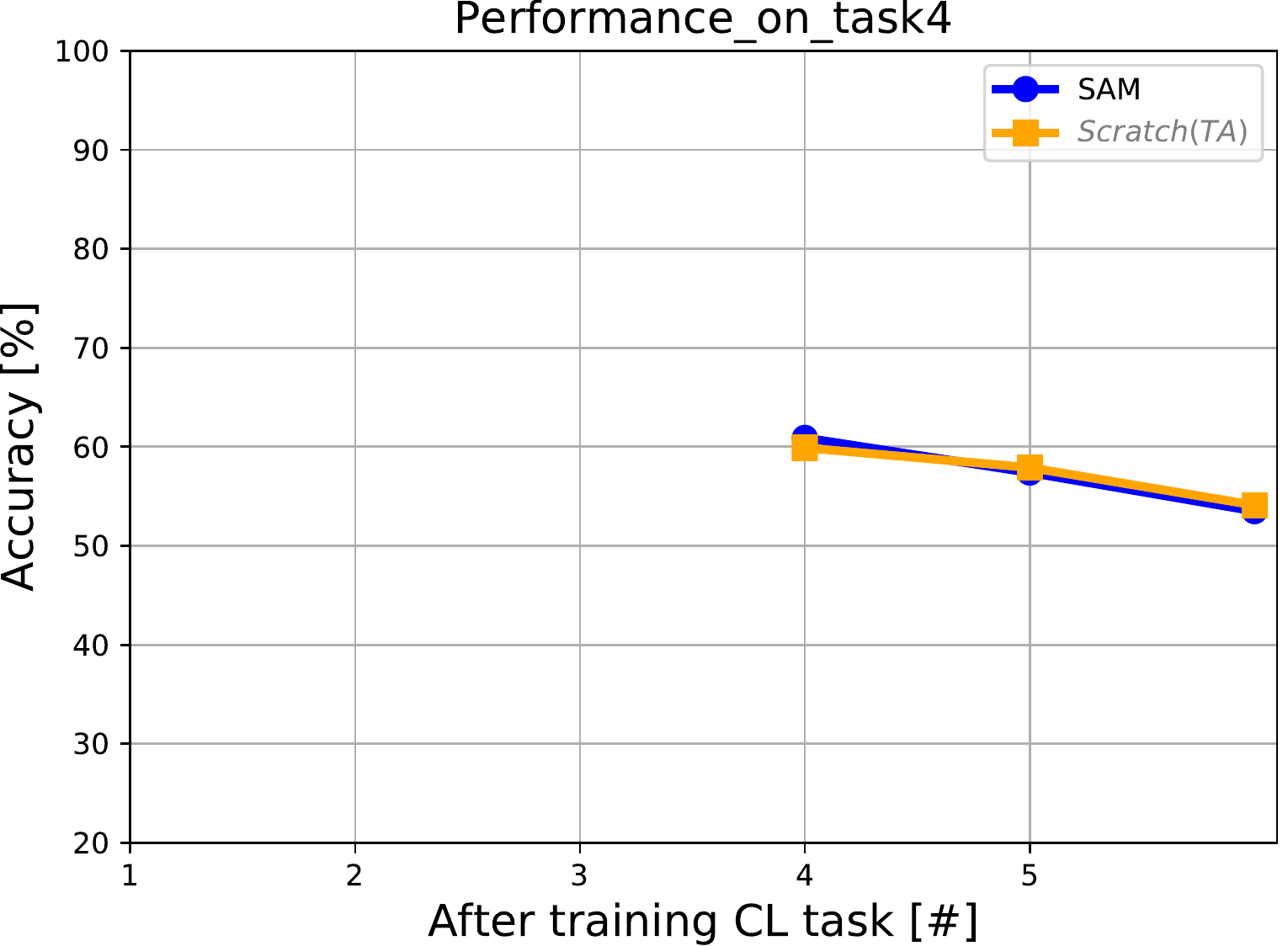}\\
  \vspace{0.4cm}
 \includegraphics[width=0.48\columnwidth]{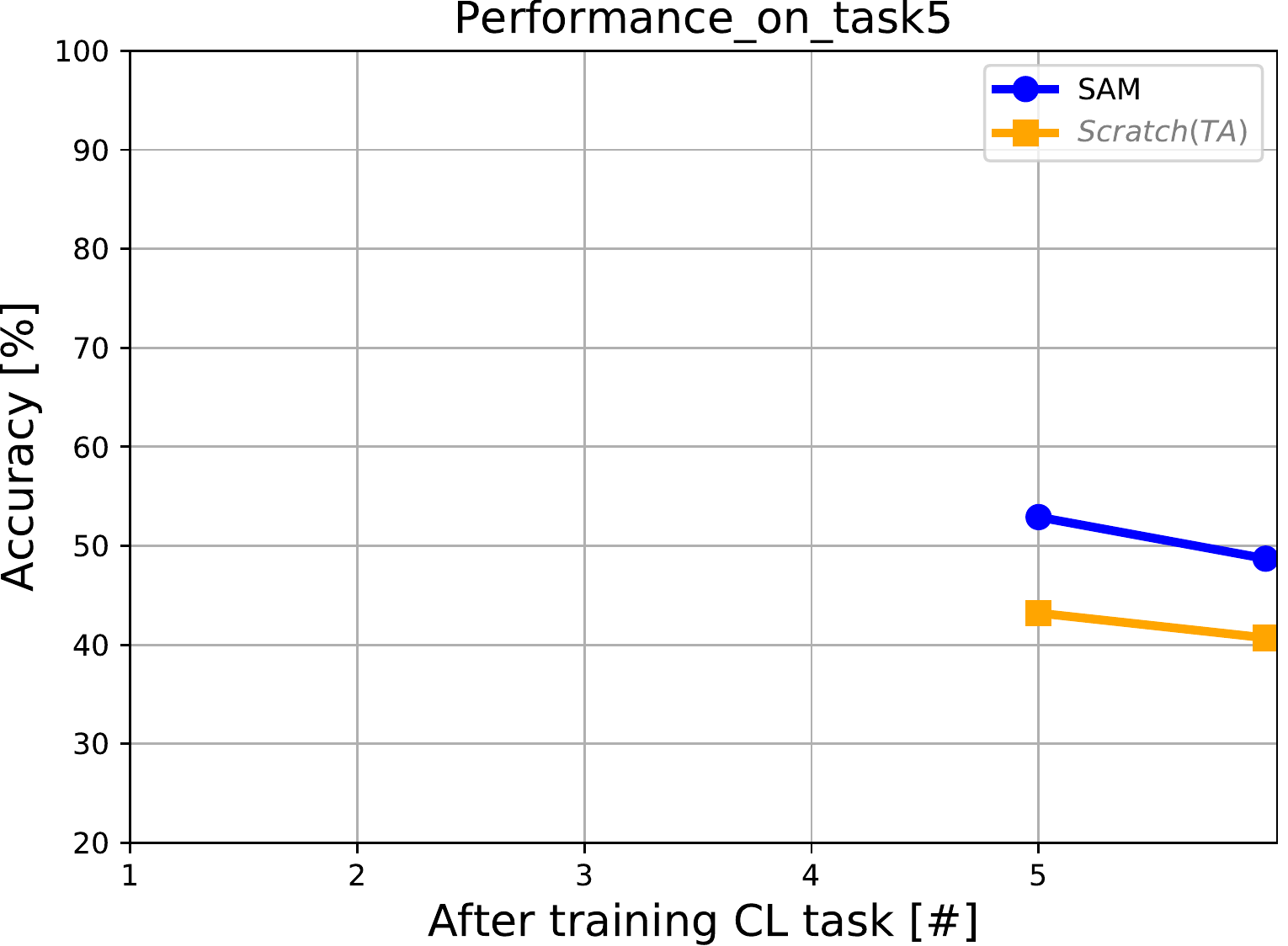}
 \includegraphics[width=0.48\columnwidth]{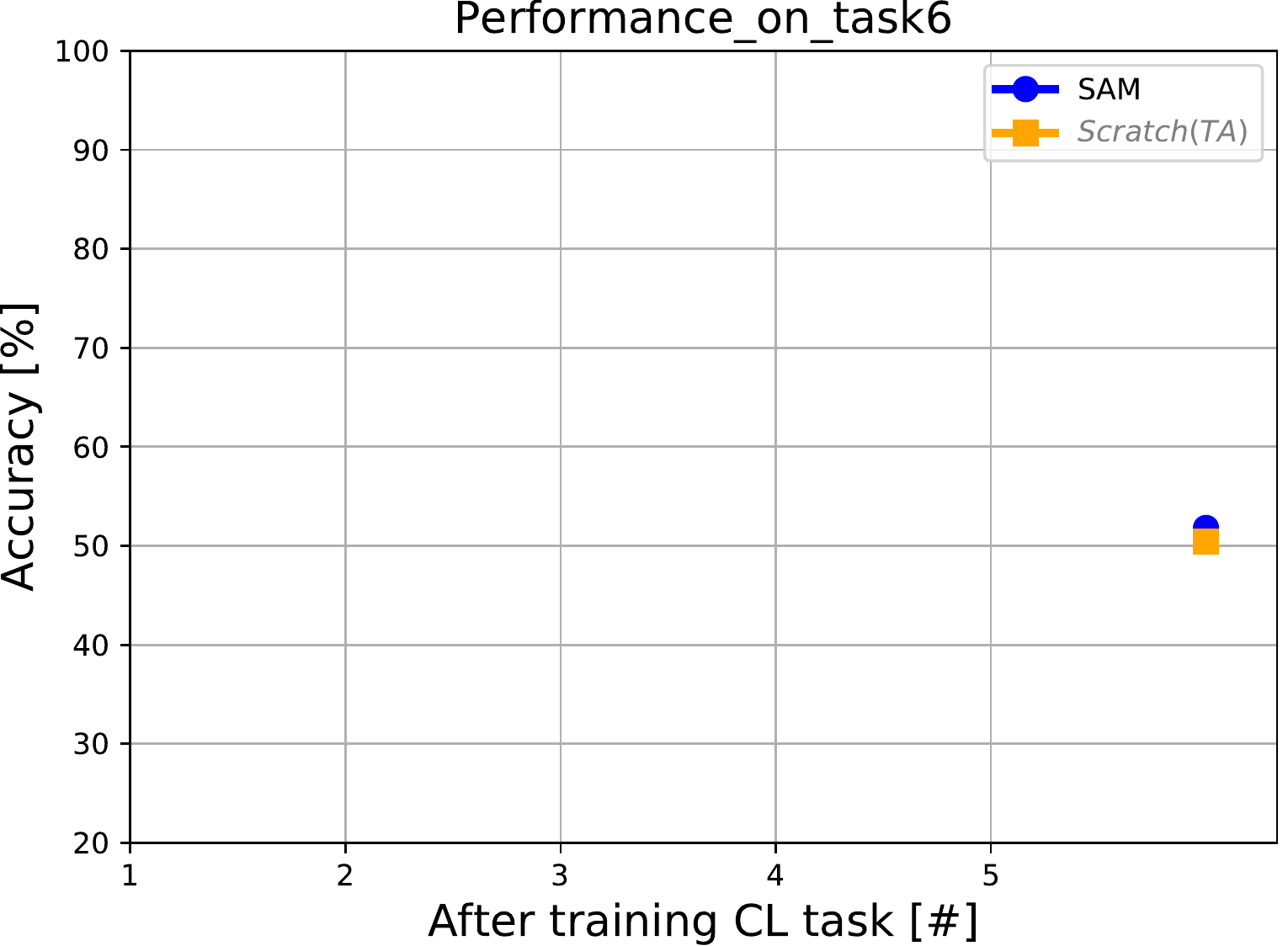}\\
 \vspace{0.4cm}
 \includegraphics[width=0.48\columnwidth]{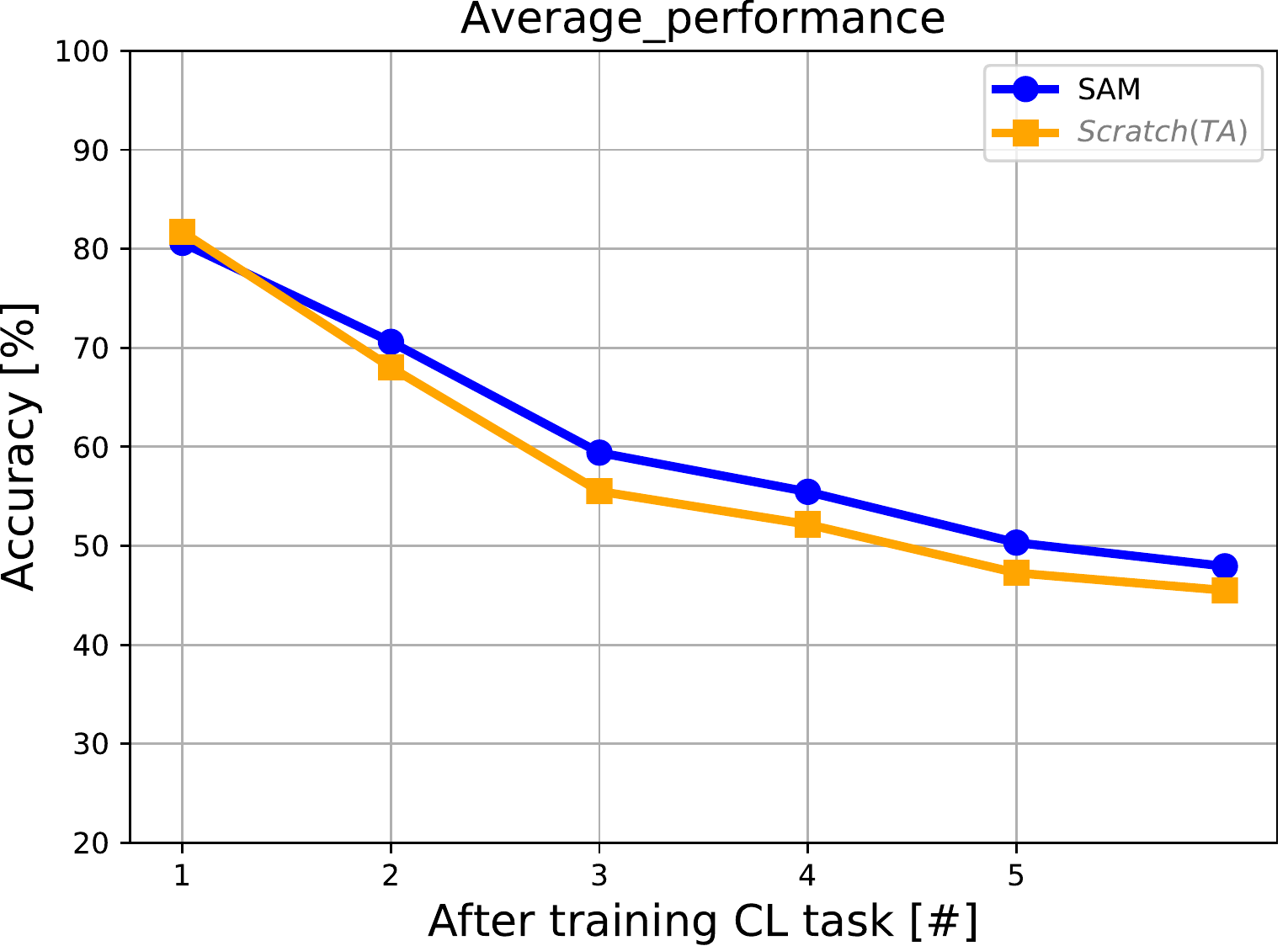}
 \caption{The first three rows show the accuracy for each task in the Split CIFAR-10/100 benchmark as a function of the number of trained tasks so far. Each panel shows the accuracy of each task starting from the point where the model faces that task and after facing each consecutive task. The last row shows the average accuracy over all the tasks learned so far.}
 \label{detailed_CIFAR10}
\end{figure}

\begin{figure}[ht]
 \centering
 \includegraphics[width=0.48\columnwidth]{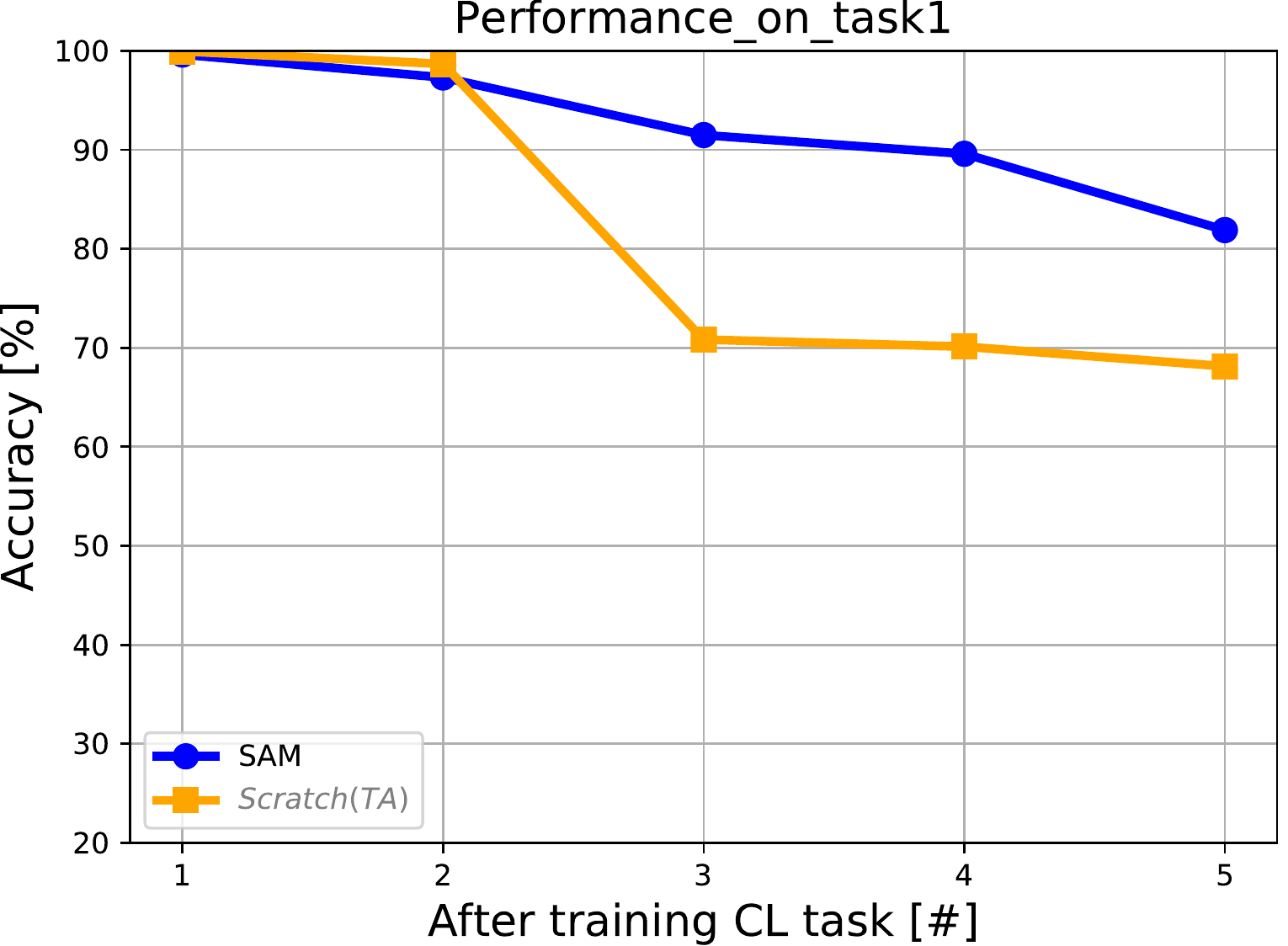}
 \includegraphics[width=0.48\columnwidth]{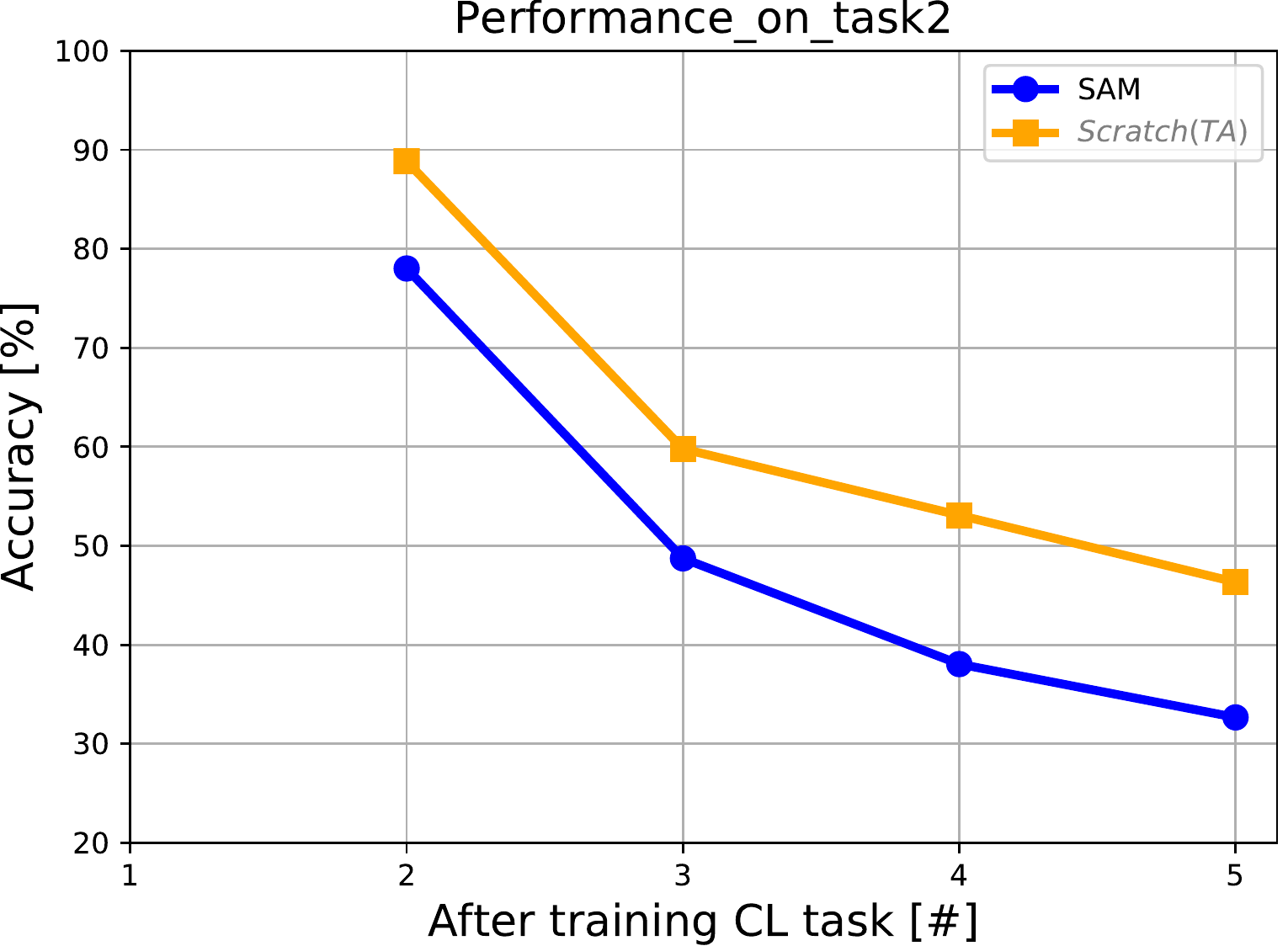}\\
 \vspace{0.4cm}
  \includegraphics[width=0.48\columnwidth]{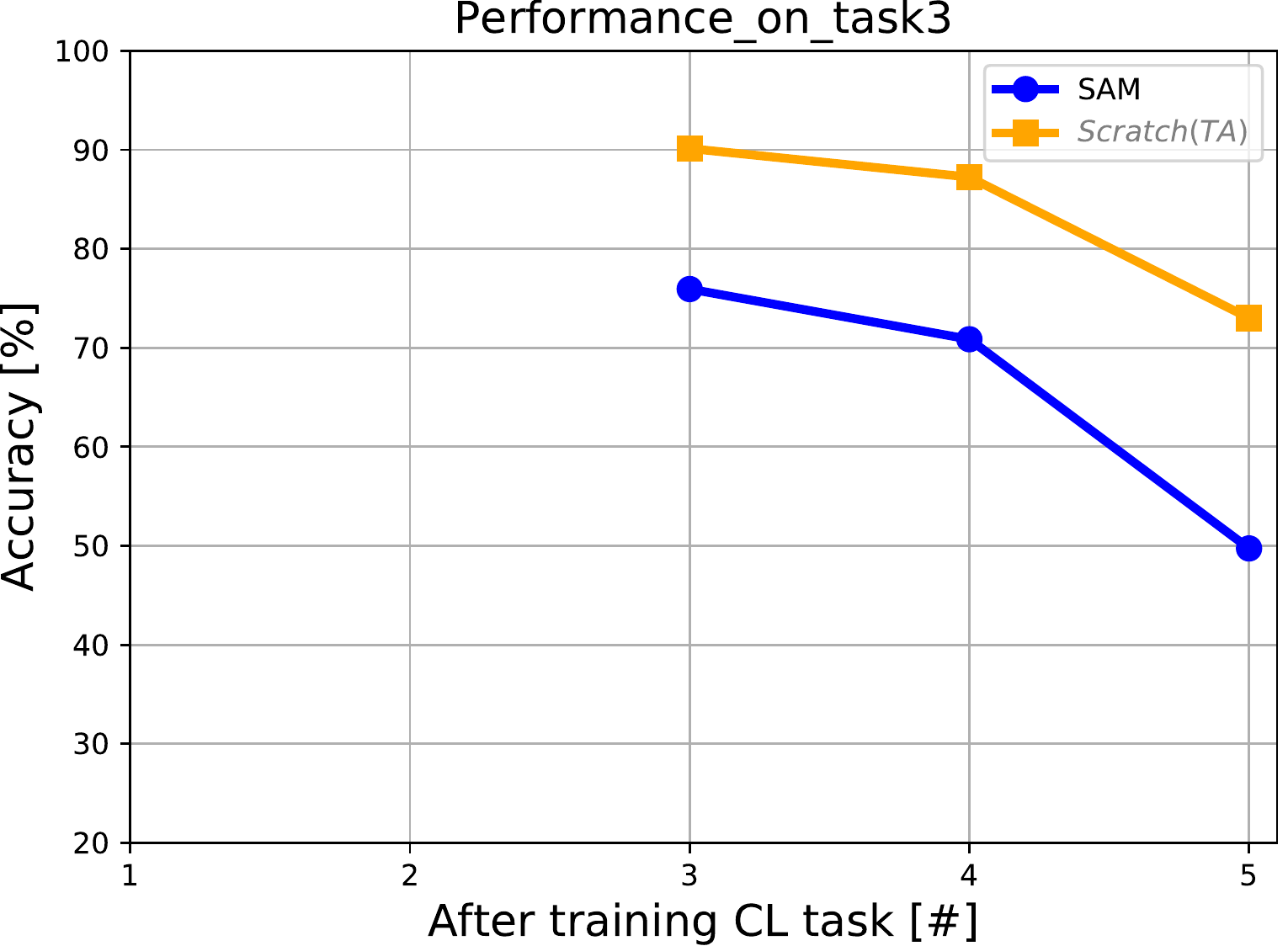}
 \includegraphics[width=0.48\columnwidth]{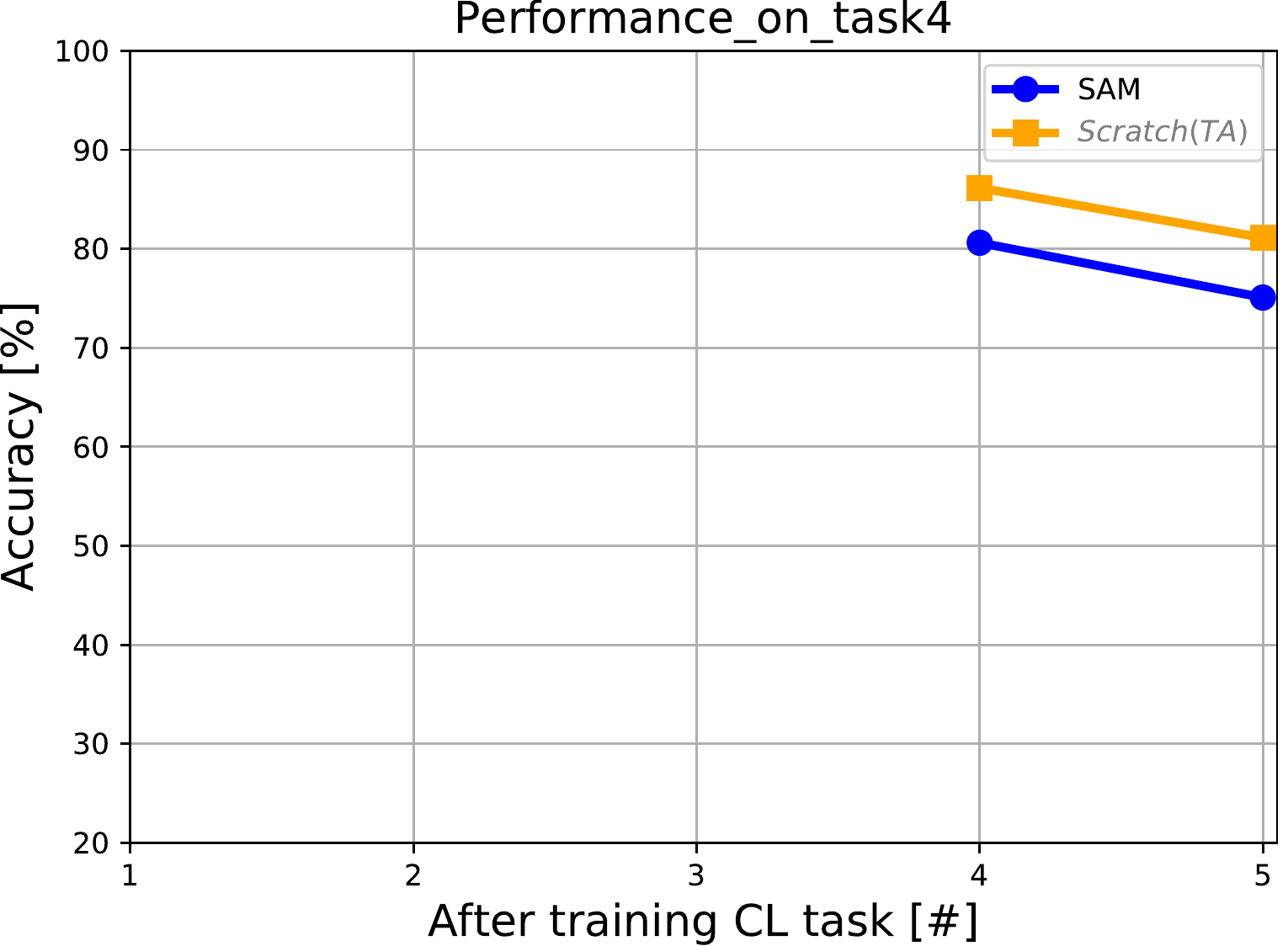}\\
 \vspace{0.4cm}
  \includegraphics[width=0.48\columnwidth]{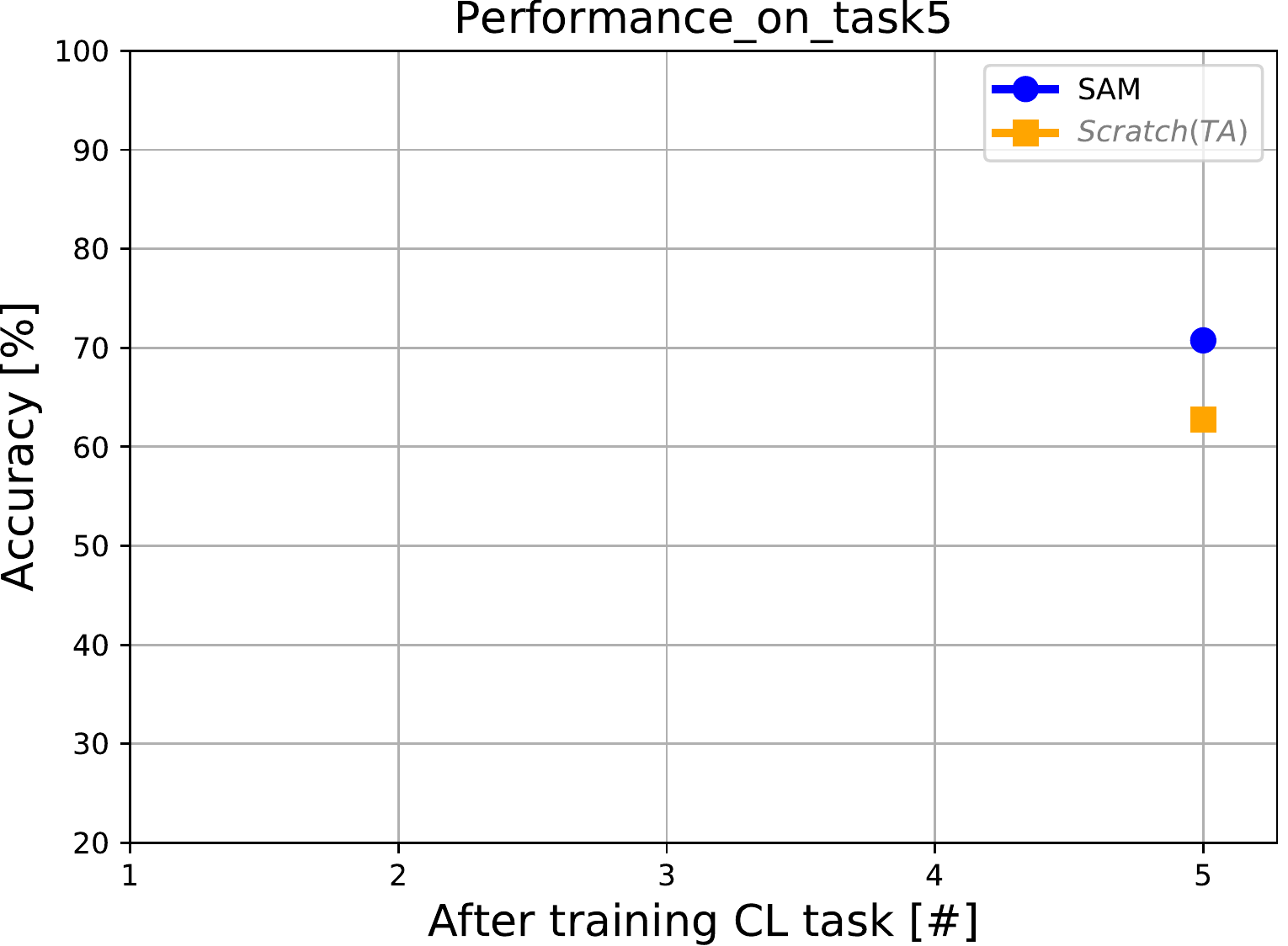}
 \includegraphics[width=0.48\columnwidth]{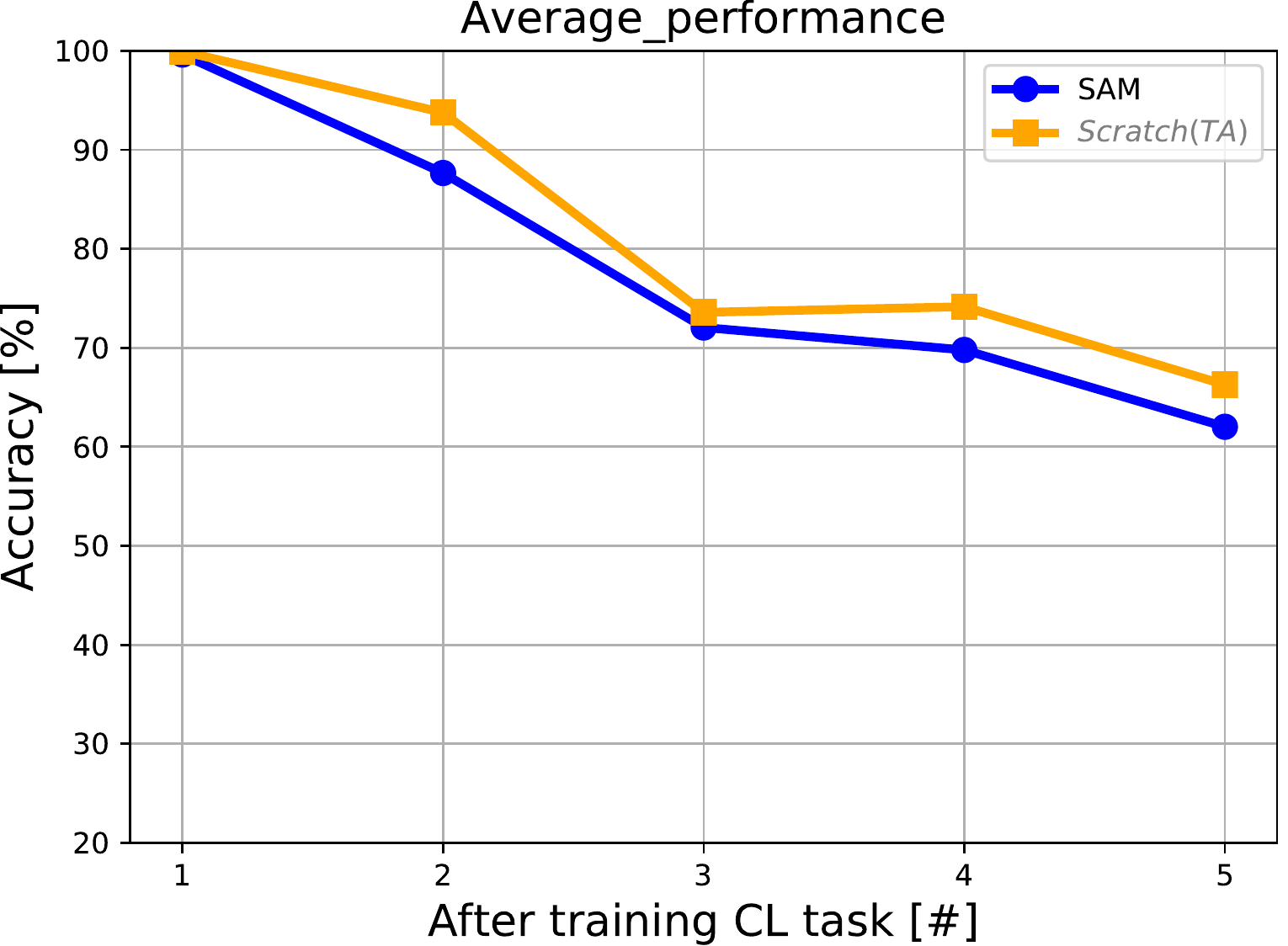}

 \caption{The first five panels show the accuracy for each task in the split MNIST benchmark as a function of the number of trained tasks so far. Each panel shows the accuracy of each task starting from the point where the model faces that task and after facing each consecutive task. The last panel shows the average accuracy over all the tasks learned so far.}
 \label{detailed_splitMNIST}
\end{figure}

In this appendix, we analyze gradually the performance of our proposed method in task agnostic scenario. We perform this analysis by computing the average accuracy on all past continual learning testing tasks after training the SAM model on each task. We also analyze the performance of each task as a function of the number of consecutive tasks. With this experiment, we would like to understand how the performance of the proposed model degrades over time, while new tasks are encountered in the training process. We compare the gradual performance of SAM to the Scratch(TA) baseline. Figure \ref{detailed_CIFAR10} and \ref{detailed_splitMNIST} show the gradual learning behavior on the Split CIFAR-10/100 and Split MNIST benchmarks respectively. This experiment allows us to dive more into the learning process of SAM. As expected the performance degrades when new tasks are encountered. It is interesting to see that SAM performs on par with the Scratch models, while actually in the Scratch baseline a new model is trained for each task from scratch. The degrade in performance for Scratch(TA) is coming from using the decision module to identify the final output in the task agnostic inference. Instead of evaluating the CL methods using the final average accuracy over all classes, this experiment would be useful for future work to determine the factors that lead to performance degradation. For instance, as shown in the figures, some tasks cause significant degradation in performance. Identifying the correct task in which the test input image belongs would help in increasing the model performance. This will open new research ideas for task agnostic inference.

\clearpage

\end{document}